%% file: ms.tex
\documentclass[sigconf]{acmart}

\usepackage[capitalize]{cleveref}

\usepackage{algorithm}
\usepackage{algpseudocode}
\usepackage[acronym,nomain]{glossaries}
\usepackage[font={small}]{caption}
\usepackage{listings}
\usepackage{enumitem}
\usepackage{subfigure}
\usepackage{todonotes}
\usepackage{xspace}
\usepackage{tabularx}
\usepackage{amsmath}
\usepackage{pgf}
\usepackage{tikz}
\usepackage{diagbox}
\usetikzlibrary{arrows,patterns}

\DeclareMathOperator{\graph}{\mathcal{G}}
\DeclareMathOperator{\vertices}{\mathcal{V}}
\DeclareMathOperator{\edges}{\mathcal{E}}
\DeclareMathOperator{\affected}{affected}
\newcommand{\neighbour}[1]{\mathcal{N}(#1)}

\renewcommand\footnotetextcopyrightpermission[1]{} % removes footnote with conference information in first column
\input{glossaries}
\makeglossaries

\begin{document}

\title{Efficient Representation Learning  Using Random Walks for Dynamic Graphs}

% Author in ACM
\author{Hooman Peiro Sajjad}
\affiliation{%
  \institution{KTH Royal Institute of Technology}
  \city{Stockholm}
  \country{Sweden}
}
\email{shps@kth.se}

\author{Andrew Docherty}
\affiliation{%
  \institution{Data61, CSIRO}
  \city{Sydney}
  \country{Australia}
}
\email{andrew.docherty@data61.csiro.au}

\author{Yuriy Tyshetskiy}
\affiliation{%
  \institution{Data61, CSIRO}
  \city{Sydney}
  \country{Australia}
}
\email{yuriy.tyshetskiy@data61.csiro.au }

% Author in IEEE

% \author{\IEEEauthorblockN{Hooman Peiro Sajjad\IEEEauthorrefmark{1}}
% \IEEEauthorblockA{\IEEEauthorrefmark{1}KTH Royal Institute of Technology\\
% Stockholm, Sweden\\shps@kth.se}}

%
% The code below should be generated by the tool at
% http://dl.acm.org/ccs.cfm
% Please copy and paste the code instead of the example below.
%
% \begin{CCSXML}
% <ccs2012>
%  <concept>
%   <concept_id>10010520.10010553.10010562</concept_id>
%   <concept_desc>Computer systems organization~Embedded systems</concept_desc>
%   <concept_significance>500</concept_significance>
%  </concept>
%  <concept>
%   <concept_id>10010520.10010575.10010755</concept_id>
%   <concept_desc>Computer systems organization~Redundancy</concept_desc>
%   <concept_significance>300</concept_significance>
%  </concept>
%  <concept>
%   <concept_id>10010520.10010553.10010554</concept_id>
%   <concept_desc>Computer systems organization~Robotics</concept_desc>
%   <concept_significance>100</concept_significance>
%  </concept>
%  <concept>
%   <concept_id>10003033.10003083.10003095</concept_id>
%   <concept_desc>Networks~Network reliability</concept_desc>
%   <concept_significance>100</concept_significance>
%  </concept>
% </ccs2012>
% \end{CCSXML}

% \ccsdesc[500]{Computer systems organization~Embedded systems}
% \ccsdesc[300]{Computer systems organization~Redundancy}
% \ccsdesc{Computer systems organization~Robotics}
% \ccsdesc[100]{Networks~Network reliability}

\input{chapters/00-abstract}


\maketitle

% \IEEEpeerreviewmaketitle

% \begin{IEEEkeywords}
% \end{IEEEkeywords}

\input{chapters/01A-introduction}
\input{chapters/02B-background}
\input{chapters/03B-problem}
\input{chapters/04-solution}

\input{chapters/06-evaluation}
\input{chapters/07-related-work}
\input{chapters/08-conclusion}
\input{chapters/09-acknowledgement}

% \bibliographystyle{IEEEtran}
\bibliographystyle{ACM-Reference-Format}
\bibliography{bibliography}

% \appendix
% \section*{APPENDIX}
% \setcounter{section}{1}

% \input{chapters/appendix}

\end{document}

%% file: glossaries.tex
\newglossaryentry{m1}
{
        name={Static},
        description={Regenerate all random walks on the current snapshot.},
}

\newglossaryentry{m2}
{
        name={Unbiased Update},
        description={Unbiased incremental random walk update.},
}

\newglossaryentry{m3}
{
        name={Fast Update},
        description={Fast biased incremental random walk update.},
}

\newglossaryentry{m4}
{
        name={Na\"{i}ve Update},
        description={Fast biased incremental random walk.},
}

\newglossaryentry{u1}
{
        name={Retrain},
        description={Ignore previous training and re-train skip-gram model.},
}

\newglossaryentry{u2}
{
        name={Incremental},
        description={Incremental update of skip-gram model.},
}

%% file: chapters/00-abstract.tex
\begin{abstract}
An important part of many machine learning workflows on graphs is vertex representation learning, i.e., learning a low-dimensional vector representation for each vertex in the graph. Recently, several powerful techniques for unsupervised representation learning have been demonstrated to give the state-of-the-art performance in downstream tasks such as vertex classification and edge prediction. These techniques rely on random walks performed on the graph in order to capture its structural properties. These structural properties are then encoded in the vector representation space.

However, most contemporary representation learning methods only apply to static graphs while real-world graphs are often dynamic and change over time. Static representation learning methods are not able to update the vector representations when the graph changes; therefore, they must re-generate the vector representations on an updated static snapshot of the graph regardless of the extent of the change in the graph. 
%Furthermore, naive updates of the random walks and representations can lead to poor performance.
In this work, we propose computationally efficient algorithms for vertex representation learning that extend random walk based methods to dynamic graphs. The computation complexity of our algorithms depends upon the extent and rate of changes\textemdash the number of edges changed per update\textemdash and on the density of the graph. We empirically evaluate our algorithms on real world datasets for downstream machine learning tasks of multi-class and multi-label vertex classification. The results show that our algorithms can achieve competitive results to the state-of-the-art methods while being computationally efficient. 

%Embedding nodes of a web-scale graph into a low-dimensional feature space is an important step for downstream machine learning tasks such as node classification. Recent research have demonstrated successful random walk based deep learning methods for feature learning, such as DeepWalk and node2vec. These methods treat graphs as documents and use the word embedding technique, the Skipgram model. Even though these methods are computationally efficient for a static graph, they cannot be used for dynamic graphs that change over time.

%Therefore, in this work, we study feature learning on dynamic graphs. We propose a set of algorithms that extends random walk and Skipgram model training for dynamic graphs. The computation complexity of our algorithms depends to the extent of the changes in a graph. We empirically evaluate our algorithms on real world data sets for multi-class and multi-label node classification. The results show that our framework can achieve competitive results to the state-of-the-art methods in terms of precision and recall while being up to two orders of magnitude faster. 

\end{abstract}

%% file: chapters/01A-introduction.tex
\section{Introduction}
Graphs or networks are the natural representation of a collection of entities and the relationships between them. They are fundamental structures that have many examples in the real world, e.g., social networks, transport networks, financial transactions, and communication networks. 
%A graph is not only useful as a structured knowledge. 
Recently, several machine learning techniques have been proposed that use graph information to predict attributes of vertices, relationships, and the entire graph \cite{hamilton2017representation}.

% Recently, there has been significant new research on machine learning techniques that explicitly use graph information to predict attributes of vertices, relationships, and entire graphs~\cite{perozzi2014deepwalk,tang2015line,grover2016node2vec,dong2017metapath2vec, hamilton2017inductive,ying2018graph,kipf2016semi}.

% Recently, there has been significant research on machine learning techniques that explicitly use graph information to predict attributes of vertices, relationships, and entire graphs~\cite{perozzi2014deepwalk,tang2015line,grover2016node2vec,dong2017metapath2vec, hamilton2017inductive,ying2018graph,kipf2016semi}

An effective approach for incorporating graph information into machine learning models is \emph{representation learning}. Representation learning seeks to learn low dimensional vector representations for the vertices (vertex embeddings). The goal of the representation learning is to find a mapping of vertices to a vector representation such that distances between these vector representations meaningfully relate to similarities in the local structure of the vertices~\cite{zhang2018review}. 
%Representation learning techniques have been extended to learn vector embeddings of edges and sub-graphs~\cite{hamilton2017representation}.

Recently, there has been significant research on representation learning and machine learning techniques that explicitly use graph information~\cite{perozzi2014deepwalk,tang2015line,grover2016node2vec,dong2017metapath2vec, hamilton2017inductive,ying2018graph,kipf2016semi}. One common approach in many of the recent methods is \emph{unsupervised} representation learning, which learns low dimensional vector representations for the vertices only based on the graph structure.

% In order to capture the structure of the graphs efficiently, random walks have been shown to be scalable to large graphs

% Unsupervised representation learning based on combining random walks with recent representation learning methods from language modelling have been shown to learn high quality representations of vertices that can be used in downstream machine learning tasks such as vertex classification and edge prediction~\cite{perozzi2014deepwalk}.

In order to capture the structure of the graphs efficiently, random walks have been shown to be scalable to large graphs~\cite{perozzi2014deepwalk}. In addition, random walks have been shown to be able to trade off structural equivalence (vertices that have similar local structure have similar embeddings) and homophily (vertices that belong to the same communities have similar embeddings)~\cite{grover2016node2vec}. Random walks are combined with recent representation learning methods from language modelling to give high quality representations of vertices that can be used in downstream machine learning tasks such as vertex classification and edge prediction~\cite{perozzi2014deepwalk, grover2016node2vec}.
In addition, random walk based methods have been extended to capture subgraph embeddings~\cite{adhikari2018sub2vec}, and vertex representations in heterogeneous graphs~\cite{dong2017metapath2vec}. Random walk based methods have been shown to have a fundamental link to matrix factorization~\cite{qiu2018network}.

However, most of the previous research on unsupervised representation learning is based on static graphs whereas most real-world graphs are dynamic \textemdash namely, they change over time. For example, in a social network, new users join the network (added vertices) and existing users add friendships (added edges).  However, the static representation learning algorithms cannot measure the extent of the change in the graph.
This presents a challenge for static and transductive representation learning algorithms such as DeepWalk~\cite{perozzi2014deepwalk} and node2vec~\cite{grover2016node2vec} applied to large dynamic graphs, in that it becomes impractical and inefficient to re-learn vertex representations from scratch upon each change in the graph.
% In addition to this, the representations learned by most random-walk based methods will change for a given vertex each time the algorithm is run.
Therefore, a method for learning representations of vertices that can explicitly utilise information about changes to the graph and update the vertex representations is desirable, in which the vertex representations can be incrementally learned as the graph changes and grows over time.

There has been recent research into how to use random-walk based representation learning for dynamic graphs. Nguyen et al.~\citep{nguyen2018dynamic} used dynamic random walks on a temporal graph where at each step the next step is restricted to edges where the time is greater than at the previous step. The representations that were learned using these dynamic random walks improved predictive performance in several downstream machine learning tasks.
Following on from this work, Winter et al.~\cite{dewinter2018combining} investigated the use of time-directed walks on dynamic graphs for edge prediction and show that the past state of the graph can be used to predict the future edges. They found that node2vec~\cite{grover2016node2vec} applied on static snapshots of the graphs perform better in some cases than using temporally-directed random walks.

In contrast to the previous works, in this work, we focus on how unsupervised learning methods based on random walks can be modified when a graph changes over time. While previous techniques have used random-walks based methods on temporal graph snapshots they have not shown how to utilise what was learned in the previous snapshot to efficiently calculate the representations in the next snapshot. We break this problem into two parts: firstly, we look at how a pre-generated set of random walks generated on one graph snapshot can be updated when the graph changes. We show that simplistic methods to perform this update give random walks that do not statistically represent the updated graph. We then propose a general random walk update algorithm that produces an updated set of walks that is statistically indistinguishable from a set of random walks generated from scratch on the new graph. Secondly, we investigate how to update vertex representations incrementally given the current set of random walks, by treating the updates as a fine-tuning step in the DeepWalk and node2vec algorithms~\cite{perozzi2014deepwalk,grover2016node2vec}.

%We show that the computation complexity of the update algorithm directly depends on the extent of changes in the graph, and is generally less than that of re-generating all random walks from scratch upon change in the graph structure. 

We demonstrate on multiple real-world datasets that our methods for updating the set of random walks and the resulting vertex representations give comparable predictive accuracy for downstream tasks to that obtained by re-learning these representations at each time snapshot of the dynamic graph while being much less expensive computationally. We discuss the trade-offs inherent in updating vertex representations, and how the computational cost of updating random walks and vertex representations depends on the number of edges that are added to the graph at each time step, and on the density of the graph.

Our contributions are as follows:
\begin{itemize}[topsep=0pt]
\item We propose an efficient algorithm that, given a graph structure change, produces an updated set of random walks statistically indistinguishable from walks generated from scratch on the updated graph. This update algorithm will be useful for any task requiring a set of random walks on a graph that is constantly changing.

\item We test this algorithm by updating the skip-gram model of \cite{perozzi2014deepwalk} to work on dynamic graphs, reducing the cost of calculating vertex representations (embeddings) by an order of magnitude compared to computing the embeddings from scratch.

\item We empirically evaluate our algorithms with several real world datasets for multi-class and multi-label classification tasks.
\end{itemize}

%% file: chapters/02B-background.tex
\section{Unsupervised representation learning and random walks}
Given an undirected and unweighted graph, $G=\{\vertices,  \edges\}$, with the set of vertices $\vertices=\{v_1, \ldots, v_{n}\}$ and edges $\edges=\{e_1, \ldots, e_{m}\}$ the goal of vertex representation learning is to determine a set of fixed length vectors, $z_k\in \mathbb{R}^{N_e}$, for each vertex $v_k$ such that similar vertices are close in the representation space.
This paper focuses on methods that learn these representations using the conditional probabilities of vertex pairs derived from random walks on the graph \cite{hamilton2017representation}. 

A typical workflow for vertex representation learning on graphs consists of the following steps: (i) update the graph, (ii) generate random walks, (iii) learn vertex representations (embeddings), (iv) train the downstream learning task (e.g., vertex classification). Steps (ii) and (iii), i.e., generating random walks and learning vertex representations, are the most resource-intensive steps in the workflow.  Step (iv) involves taking the learned vertex representations and using them as features for predictive tasks such as vertex classification or edge prediction. The vertex representations are learned separately using an unsupervised objective function as this gives multi-purpose vertex representations and also allows the use of a small number of labelled vertices in the downstream predication task~\cite{perozzi2014deepwalk}.

In this section, we describe how the DeepWalk and node2vec algorithms \cite{perozzi2014deepwalk, grover2016node2vec} on a static graph use random walks to generate vertex pairs that are then used to find the vector representations $z_k$ for all vertices in the graph. In the following section, we will discuss the issues that arise when applying these algorithms to streaming graphs.

\subsection{Random Walks}
In general, a random walk can be modelled as $k$-th order Markov chain, in which the state space is the set of graph vertices $V$ and the future state depends on the last $k$ steps. For a $k$th-order walk the transition probability only depends upon the previous $k$ vertices visited by the walk~\cite{benson2017spacey}.

A random walk of length $L$ starting from a vertex $v_{w(0)}$ consists of a sequence $W[w(0)] = \{ w(0), \ldots, w(L-1) \}$,
% \begin{equation*}
% W[w(0)] = \{ w(0), \ldots, w(L-1) \}
% \end{equation*}
where $w(i) \in \{0, 1, \ldots, n\}$ represents the vertex index at the $i$-th position in the walk. In general, a $k$-th order random walk is generated by sampling a vertex $w(i)$ given the $k$ previous vertices $w(i-k), \ldots, w(i-1)$ from the transition probability distribution:
\begin{equation}
p(w(i) | w(i-1), \ldots, w(i-k)),
\label{eq:rw-k}
\end{equation}
which is non-zero only if there is an edge between vertices $v_{w(i-1)}$ and $v_{w(i)}$. To generate a $k$-th order walk, we must also sample the $k$ initial vertices from another random sampling method in order to calculate the transition probability (\cref{eq:rw-k}). For first order walks this amounts to sampling the vertices to start the walk from, for higher order walks we also need another lower-order walk process to generate the $k$ initial vertices.

To make things concrete, in the next sections we will only discuss first-order and second-order random walks that are used in DeepWalk~\cite{perozzi2014deepwalk} and node2vec~\cite{grover2016node2vec} respectively.

\subsection{Vertex Pairs from Random Walks}\label{sec:vertex-pairs}
Given a set of random walks, we extract vertex pairs that appear close to each other in the walk. These vertex pairs represent the structure of the graph and will be used in the next subsection to learn the vertex representations.

Consider the case where we have a set of $K$ random walks  $W_{\graph} = \{ W[w_j(0)], \forall j \in \{1, \ldots, K\} \}$, that start from a set of initial vertices $V_0 = \{w_1(0), w_2(0), \ldots, w_{K}(0)\}$ in the graph, where $W[w_j(0)] = \{w_{j}(0), \ldots, w_{j}(L-1) \}$ is an independently created random walk starting from the vertex $w_j(0)$.

Given such a set of random walks, we sample vertex pairs from each random walk in a way analogous to words being sampled from sentences in the skip-gram model of \cite{mikolov2013efficient}. Namely, for each random walk $W = \{ w(0), \ldots, w(L-1) \}$ and for each vertex $w(i)$ we take the $p$ vertices before $w(i)$ in the walk and create $p$ pairs, giving the following set:
\begin{equation}
S^-_{p}(j,l) = \{ \bigl(w_{j}(l), w_{j}(l-p)\bigr), \ldots, \bigl(w_{j}(l), w_{j}(l-1)\bigr) \},
\end{equation}
the same is done for the $p$ vertices after the vertex $w(i)$, giving the following set of pairs:
\begin{equation}
S^+_p(j,l) = \{ \bigl(w_{j}(l), w_{j}(l+1)\bigr), \ldots, \bigl(w_{j}(l), w_{j}(l+p)\bigr) \}.
\end{equation}
Following the terminology of \cite{mikolov2013efficient} we will call the first item in the pairs
the \emph{target} and the second item the \emph{context}. Parameter $p$ is called the \emph{context window size}.

The generated target-context pairs for all random walks in set $W_G$, together they form the corpus $C_{p}(W_G)$ of vertex pairs. We define the set of all vertex pairs generated from the items before the target vertices as:
\begin{equation}
    C^-_p = \bigcup_{\substack{0 \leq j < K}} \bigcup_{0 \leq l < L} S^-_{\min(l,p)}(j, l),
\label{eq:C-}
\end{equation}
and all the pairs generated from items after the target vertices as:
\begin{equation}
    C^+_p = \bigcup_{\substack{0 \leq j < K}} \bigcup_{0 \leq l < L} S^+_{\min(L-l-1,p)}(j, l),
\label{eq:C+}
\end{equation}
finally, the complete vertex pair corpus is given by the union of both of these:
\begin{equation}
    C_{p}(W_G) = C^+_{p} \cup C^-_{p}
\label{eq:Call}
\end{equation}
, which are used to optimise the loss function as discussed in the next section.

\subsection{The Skip-Gram Model}
Given a corpus of target-context pairs $C_p(W_G)$, the vertex representations are found by learning a vector representation for each vertex that when combined by a specified function approximates the probability of co-occurrence of the target and context vertices in the vertex pair corpus. In particular, the skip-gram model of \citep{mikolov2013efficient} models the conditional probability of a vertex pair,
$(v_t, v_c)\in \vertices \times \vertices$,  by a log-linear function of the inner product between the vectors $z_t$ and $z_c$ representing the vertices, as follows:
\begin{equation}
p(v_t|v_c) \simeq \frac{\exp(z_t \cdot z_c)} {\sum_{v_k\in \vertices} \exp(z_t \cdot z_k)}
\label{eq:cond_prob}
\end{equation}
The vertex representation vectors $z_k$ for all vertices  $v_k\in\vertices$ can be found by minimising the following cross-entropy loss function:
\begin{equation}
\sum_{(v_t, v_c) \in \mathcal{C}_p(W_G)} \log\left( \frac{\exp(z_t \cdot z_c)} {\sum_{v_k\in \vertices} \exp(z_t \cdot z_k)} \right)
\label{eq:unsup-obj}
\end{equation}

As the partition function in the denominator of Eq.~\ref{eq:unsup-obj} is a sum over all vertices, it is computationally intractable for all but the smallest graphs. Therefore, more efficient formulations are used to approximate this formulation, in particular Perozzi et al.~\cite{perozzi2014deepwalk} use a hierarchical softmax to approximate the partition function. In this work, as in more recent works~\cite{grover2016node2vec, hamilton2017inductive}, we use negative sampling~\cite{mikolov2013distributed} to estimate Eq.~\ref{eq:unsup-obj}, which is more efficient compared to hierarchical softmax. 

\subsection{Random Walk Length}

Grover et al.~\citep{grover2016node2vec} found that walks of length $80$ gave the best cross-validated performance on a downstream vertex classification task when all other hyper-parameters are fixed. This meant that the number of walks was fixed and as the walk length is increased the total number of target-context pairs in the corpus  $C_p(W_{\graph})$ also increases.

In contrast, we claim that shorter walks can give embeddings that have similar performance results on a downstream task when the number of walks is adjusted to keep the training corpus size the same.
\cref{table:wl-vs-nw} shows similar performance of a downstream vertex classification task on the embeddings calculated from random walks of lengths $l=10$ and $l=80$.
The performance is measured as the Macro-F1 scores for the multi-class classification on the Cora and CoCit datasets, where $r$ is the number of walks starting from each vertex and $l$ is the walk length. 
We keep the size of the corpus $C_p(W_G)$ the same in the experiments by increasing the number of walks per vertex when we decrease the length of the walks.
More information about the datasets and the experimental setups are given in \cref{sec:eval}.

However, we note that the training corpus $C_p(W_G)$ will be affected by the length of the walks used to generate it.  
Specifically, for first-order random walks, the length of the walks will firstly change the unigram distribution of vertices in the training corpus, and secondly change the bi-gram distribution of vertex pairs through edge effects. 

Firstly, the unigram distribution of the vertices appearing in the corpus will change with random walk length. In particular, as the length of the random walks becomes long, the singleton distribution of vertices will tend to the stationary distribution~\cite{bollobas1998graphtheory}. In comparison, short random walks will be dominated by the distribution of the initial vertices, typically a uniform distribution.

The unigram vertex distribution effectively alters the overall weighting of probabilities for each vertex in the cross-entropy loss function \cref{eq:unsup-obj}. The effect is for longer walks to give a comparatively higher weight to high-degree vertices, as they will appear more frequently in longer walks.

Secondly, edge effects due to the sampling method of \cref{eq:C-} and  \cref{eq:C+} will bias the vertex pair corpus $C_p(W_G)$ towards vertices that are closer together. This is because when the target vertex is closer than $p$ to the start of the walk not all $p$ vertex pairs can be sampled by \cref{eq:C-}. Similarly, \cref{eq:C+} is biased to vertex pairs that are closer together at the end of the walk. 
As the length of the walk increases, the number of vertex pairs sampled with the full context window will increase and the effect of the edges will proportionally decrease.
Therefore, the corpus $C_p(W_G)$ will have a larger bias towards vertices that are closer together for short walks compared to long walks.
For higher-order walks, there are similar edge effects due to the choice of the initial $k-1$ nodes affecting the vertex pair distribution more for shorter walks.

In this section, we have described two ways that the length affects the generation of vertex pairs from random walks. We show that the effects of walk length are minimal when the size of the vertex pair corpus is controlled for. Furthermore, the changes to the vertex pair corpus caused by different walk lengths can potentially be controlled for in other ways, for example by changing the distribution of the initial vertices of the random walks.

\begin{table}[t]
\centering
\caption{Multi-class classification on the Cora and Cocit datasets for different walk lengths ($l$) and number of walks ($r$).}
\scalebox{1}{
\begin{tabular}{ c|cc }
 Configuration & Cora & CoCit\\
 \hline
 $r=80,l=10$ & 0.7825 & 0.3143\\
 $r=10,l=80$ & 0.7844 & 0.3059\\
\end{tabular}}
\vspace{-1.5em}
\label{table:wl-vs-nw}
\end{table}

%% file: chapters/03B-problem.tex
\section{Dynamic Graphs}\label{sec:dynamic-networks}
A dynamic graph can be represented as a series of undirected and unweighted graphs, $\graph^t=\{\vertices^t, \edges^t\}$, where vertices $\vertices^t=\{v_1^{t},...,v_{n(t)}^{t}\}$, edges $\edges^t=\{e_1^{t},...,e_{m(t)}^{t}\}$, and $t$ is a discrete series of times. A dynamic graph can be considered as a set of updates taking the graph at time $t$ and producing a modified graph at time $t+1$. These updates consist of deleting or adding one or more vertices and edges. We represent the set of vertices and edges that are deleted from the graph between times $t$ and $t+1$ as $D_{\vertices}^{t+1}$ and $D_{\edges}^{t+1}$, and the set of vertices and edges that are added to the graph between times $t$ and $t+1$ as $A_{\vertices}^{t+1}$ and $A_{\edges}^{t+1}$ respectively. Therefore, the updated graph at time $t$ can be given in terms of the vertices and edges as $\vertices^{t+1} = (\vertices^{t} \setminus D_{\vertices}^{t+1}) \cup A_{\vertices}^{t+1}$ and $\edges^{t+1} = (\edges^{t} \setminus D_{\edges}^{t+1}) \cup A_{\edges}^{t+1}$.
% \begin{align*}
% \vertices^{t+1} &= (\vertices^{t} \setminus D_{\vertices}^{t+1}) \cup A_{\vertices}^{t+1} \\
% \edges^{t+1} &= (\edges^{t} \setminus D_{\edges}^{t+1}) \cup A_{\edges}^{t+1}.
% \end{align*}

Our aim is given a corpus of random walks on a graph at time $t$, $W_{G^t}$, to update the random walks so that the updated set at time $t+1$,  $W_{G^{t+1}}$, is statistically representative of the updated graph. Namely, the updated corpus of random walks at time $t+1$ should be statistically the same as random walks drawn only from the current graph snapshot $G^{t+1}$.

Now, when updating random walks on graphs we consider the set of vertices which have changed directly as a result of the additions and deletions of vertices and edges. 
Specifically we denote the set of vertices contained in all added edges as $\vertices\bigl(A_{\edges}^{t+1}\bigr)$ and the set of vertices contained in all removed edges as $\vertices\bigl(D_{\edges}^{t+1}\bigr)$.
There are several different cases to consider:

\begin{itemize}
\item Deleted vertices with edges in $D_{\vertices}^{t+1} \cap \vertices\bigl(D_{\edges}^{t+1}\bigr)$ will be removed from the graph and any random walks containing these vertices will be invalid.

\item Deleted vertices without edges in $D_{\vertices}^{t+1} \setminus \vertices\bigl(D_{\edges}^{t+1}\bigr)$ will not be in any valid random walks and no change to the random walk corpus is needed.

\item Added vertices without edges $A_{\vertices}^{t+1} \setminus \vertices\bigl(A_{\edges}^{t+1}\bigr)$ will not be included in any random walks from other vertices.

\item All other added edges and deleted edges without removed vertices affect all random walks that include the vertices connected by these edges.
\end{itemize}

To analyse the effect of graph changes on the random walks we define the following terms:
\begin{itemize}
\item \emph{Affected vertices}: all vertices that are in the set of vertices contained in the set of added edges, $\vertices\bigl(A_{\edges}^{t+1}\bigr)$ and the set of removed edges $\vertices\bigl(D_{\edges}^{t+1}\bigr)$ but without the vertices in the deleted vertex set:
\begin{equation}\label{eq:affected}
    \vertices^{t+1}_{\affected} = \vertices\bigl(A_{\edges}^{t+1}\bigr) \cup \vertices\bigl(D_{\edges}^{t+1}\bigr) \setminus D_{\vertices}^{t+1}
\end{equation}

\item \emph{Affected walks}: All random walks from the corpus of random walks $W_{G^{t}}$ that contain at least one affected vertex.
\end{itemize}

Importantly, we note that all un-affected walks on the graph represent valid samples of walks in the current graph after the update, $\graph^{t+1}$. On the other hand, the affected walks do not represent the statistics of the current graph. At the point that an affected walk encountered an affected vertex the next step of the walk would have different transition probabilities on the current graph snapshot, $\graph^{t+1}$, than the previous graph snapshot, $\graph^{t}$. Therefore, the first encountered affected vertex in an affected walk is of special importance. 
This generalises to all random walk orders, as it is only when the random walks pass through an affected vertex that different transition probabilities of next step of the walk are of consequence.

\subsection{Updating the Random Walk Corpus}\label{subsec:dyanamic-rw-problem}
Our goal is to update the random walk corpus so that it is indistinguishable from random walks generated on the updated graph. Therefore, the baseline that we will compare to is to re-generate random walks for the latest snapshot of the graph every time the graph is changed. This baseline random walk algorithm, the \emph{\gls{m1} algorithm}, is given in pseudocode in \cref{alg:m1}.

% I'm not sure about the next paragraph
The input data for the \gls{m1} algorithm is the snapshot of the graph $\graph^{t+1}$, and the random walk parameters: number of walks $r$ and walk length $l$. \gls{m1} takes all the vertices of the snapshot $\graph^{t+1}$ and initialises $r$ random walks per vertex (Line~\ref{alg:l:m1-init}), with that vertex as the initial step of those $r$ random walks. The initialised walks are given to the function \emph{randomwalk} (Line~\ref{alg:l:m1-rw}) that executes random walks of length $l$ through sampling vertices from \cref{eq:rw-k}.

As this baseline algorithm requires a large amount of computation at each update, regardless of how small the numbers of added or deleted vertices and edges are, clearly a more efficient algorithm is needed. Such an algorithm would replace the minimum number of random walks in the current corpus of random walks $W_{G^t}$ with new random walks such that the updated random walk corpus, $W_{G^{t+1}}$, is statistically representative of the new graph $G^{t+1}$.

% Firstly, we consider a naive algorithm where we generate random walks for only the new vertices added to the new graph $G^{t+1}$ discard the random walks for all affected vertices in the graph $G^t$ and generate new random walks for all affected vertices $\vertices^{t+1}_{\affected}$. This random walk update algorithm, the \emph{\gls{m4}} algorithm, is given in pseudocode in \cref{alg:m4}.

Firstly in the next section, we show the problem with naively updating random walks by appealing to a simple example and we introduce the \gls{m4} algorithm. 

\subsection{Example Naive Random Walk Update}
\begin{figure}
\begin{minipage}[t]{.5\linewidth}
\centering
(a)\\
\begin{tikzpicture}[node distance=15mm, line width=1pt,
c/.style={circle, fill=white, draw=black, text=black},
aff/.style={circle, pattern=north west lines, fill=white, draw=dashed, text=black}
]
  \node (A) [c] {A};
  \node (B) [c, right of=A, yshift=5mm] {B};
  \node (C) [c, right of=B, yshift=-5mm] {C};

  \path (A) edge[-] (B)
        (B) edge[-] (C);
\end{tikzpicture}
\end{minipage}%
\begin{minipage}[t]{.5\linewidth}
\centering
(b)\\
\begin{tikzpicture}[node distance=15mm, line width=1pt,
c/.style={circle, fill=white, draw=black, text=black},
aff/.style={circle, fill=red!20, draw=black, text=black}
]
  \node (A) [c] {A};
  \node (B) [aff, right of=A, yshift=5mm] {B};
  \node (C) [c, right of=B, yshift=-5mm] {C};
  \node (D) [aff, below of=B] {D};

  \path (A) edge[-] (B)
        (B) edge[-] (C)
        (B) edge[dashed] (D);
\end{tikzpicture}
\end{minipage}
\caption{(a) An example graph $\graph^{1}$, (b) an example $\graph^{2}$, with an added vertex D and edge B$\rightarrow$D.}
\vspace{-1.5em}
\label{fig:graph_1_2}
\end{figure}
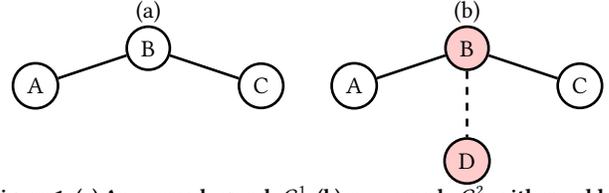

To illustrate how a naive algorithm produces biased random walks we consider taking first-order random walks of length 3 starting from a vertex uniformly sampled from the graph $\graph^1$ of \cref{fig:graph_1_2}a.
The set of walks generated will be uniformly distributed over the walks shown in \cref{table:ex_random_walks_1_2}(a).
Next, we generate the corpus of pairs $C^+_1$ as described in \cref{sec:vertex-pairs}.
This set of adjacent pairs will be uniformly distributed over the pairs shown in  \cref{table:ex_random_walks_1_2}(b).

The frequency of the vertex pairs $C_{1}^+$ can be directly related to the transition probabilities for the vertex pairs given by \cref{eq:rw-k} with $k=1$, namely in the expectation:
\begin{equation}\label{eq:empirical_trans_probs}
    E\biggl[\frac{|(t,c)|}{|(t,-)|}\biggr] = \frac{1}{|\neighbour{v_t}|},
\end{equation}
where $|(t,c)|$ is the number of pairs $(v_t, v_c)$ in the corpus $C^+_1$ and $|(t, -)|$ is the number of all pairs containing the vertex $v_t$ in the first position.
For the graph $G^1$ the expected pair probabilities \cref{eq:empirical_trans_probs} are
given by: $p(A | B) \rightarrow \frac{1}{2}$, and $p(C | B) \rightarrow \frac{1}{2}$.

Now consider the graph $\graph^2$ shown in \cref{fig:graph_1_2}b which contains the same three vertices -- A, B, and C -- as $\graph^1$ but adds a new vertex D and an edge B$\rightarrow$D. The set of affected vertices are \{B,D\} and are shown as shaded in \cref{fig:graph_1_2}b.
To update the walks to reflect the new vertex, we will naively generate walks of length 3 from the vertex D, of the same number as the expected number of walks from the other three vertices. Doing this we generate random walks that will be uniformly distributed over those shown in \cref{table:ex_random_walks_1_2}(c).
However, the corpus of pairs for the set of walks consisting of the original walks of \cref{table:ex_random_walks_1_2}(a) and the walks of \cref{table:ex_random_walks_1_2}(c) do not represent the graph statistics. The vertex pairs generated from these walks will follow the distribution shown over the pairs shown in \cref{table:ex_random_walks_1_2}(d).

Theoretically the transition probability for vertices A, C, and D given that the random walk is at vertex B should all be $1/3$ as there are now three neighbours for that vertex. However, using \cref{eq:empirical_trans_probs} to calculate the expected empirical conditional probabilities for the vertex pairs we obtain highly biased probability estimates: $p(A | B) \rightarrow 0.4\dot{4}$, $p(C | B) \rightarrow 0.4\dot{4}$, and $p(D | B) \rightarrow 0.1\dot{1}$.

\begin{table}[t]
\caption{(a) Random walks for graph $\graph^1$, (b) vertex pairs for the random walks of (a), (c) new walks for graph $\graph^2$, (d) vertex pairs for the random walks of (a) and (c).}
\scalebox{0.8}{
\hspace{-6em}
\begin{minipage}[t]{.27\linewidth}
\centering
(a)\\
\begin{tabular}{ccc}
 \hline
 \multicolumn{3}{c}{Random walk} \\
 \hline
 A & B & C \\
 A & B & A \\
 B & A & B \\
 B & C & B \\
 C & B & C \\
 C & B & A \\
 \hline
\end{tabular}
\end{minipage}%
\begin{minipage}[t]{.3\linewidth}
\centering
(b)\\
\begin{tabular}{cc} 
 \hline
 Pairs $C^+_1$ & Freqency \\
 \hline
 (A, B) & 1/4 \\
 (B, C) & 1/4 \\
 (B, A) & 1/4 \\
 (C, B) & 1/4 \\
\end{tabular}
\end{minipage}
\hspace{1em}
\begin{minipage}[t]{.27\linewidth}
\centering
(c)\\
\begin{tabular}{ccc}
 \hline
 \multicolumn{3}{c}{Random walk} \\
 \hline
 D & B & C \\
 D & B & A \\
 D & B & D \\
\end{tabular}
\end{minipage}%
\begin{minipage}[t]{.25\linewidth}
\centering
(d)\\
\begin{tabular}{cc} 
 \hline
 Pairs $\tilde{C}^+_1$ & Freqency \\
 \hline
 (A, B) & 3/18 \\
 (B, C) & 4/18 \\
 (B, A) & 4/18 \\
 (C, B) & 3/18 \\
 (D, B) & 3/18 \\
 (B, D) & 1/18 \\
 \hline
\end{tabular}
\end{minipage}
}
\hspace{-3em}
\vspace{-1em}
\label{table:ex_random_walks_1_2}
\end{table}

We see that simply adding random walks generated on the newly added vertices in the updated graph $\graph^2$ to the walks generated on the old graph $\graph^1$ gives \emph{biased} conditional probability distributions and counts of vertex pairs that do not reflect the statistics of the updated graph. As the naive incremental random walk, we introduce a slightly more sophisticated algorithm, namely \gls{m4} algorithm (see \cref{alg:m4}), which initialises random walks for the affected vertices and creates random walks of length $l$. In the end, it updates the random walks $W^{t}$ by replacing the old walks by their corresponding re-generated walks and adding the new random walks for the new vertices (Line~\ref{alg:l:m4-update} in \cref{alg:m4}).

However, as explained in the example, the \gls{m4} algorithm does not give the same statistics of the random walks as rerunning the static algorithm \gls{m1} on the current state of the graph $G^{t+1}$. In other words, the random walks given by the \gls{m4} algorithm have biased empirical transition probabilities and thus do not represent the statistic structure of graph $G^{t+1}$.

In the next section, we present an algorithm that updates a set of random walks such that all statistics derived from it are identical to generating random walks on the new graph from all vertices.

%% file: chapters/04-solution.tex
\section{Dynamic Representation Learning Algorithms}
In this section, we introduce an algorithm that updates the random walks on a dynamic graph to maintain the statistical properties of the random walks as compared to sampling the random walks from scratch on the current static snapshot of the graph. We also discuss how the vertex representations learned by the skip-gram model should be updated given the vertex representations of the previous state of the graph, and the updated random walk corpus.

\subsection{Unbiased Random Walk Updates}\label{subsec:dyanamic-rw}
In the previous section, we showed that the baseline \gls{m1} algorithm, namely re-generating all walks on every graph change regardless of the extent of the change, is inefficient and incurs unnecessary computation.  We additionally showed that the \gls{m4} algorithm produces a random walk corpus that does not match the statistical properties of the updated graph. 

\input{chapters/algorithms/m1}

\input{chapters/algorithms/m4}
\input{chapters/algorithms/m2}

To motivate an algorithm that allows an update to our random walk corpus $W_{G^t}$, we return to the definitions of affected vertices and affected walks given in \cref{sec:dynamic-networks} and consider a first-order random walk. In order to update a corpus of random walks we consider a random walk that has arrived at a vertex $v$, if this vertex is not \emph{affected} there is no change to the neighbours of this vertex; therefore, the choice of next vertex in the random walk is unchanged. However, if the vertex $v$ is affected $v\in\vertices^{t+1}_{\affected}$ then the neighbours of the vertex have changed and the random walk is biased from this point as it has not considered the correct transition probabilities at the affected vertices.  A corollary is that only random walks that contain affected vertices, namely the affected walks, need to be updated in the random walk corpus. 

% To correct for this 
We propose the \gls{m2} algorithm, which is represented in \cref{alg:m2}. The \gls{m2} algorithm updates only the affected walks by re-sampling these walks starting from the first appearance of the affected vertex. Specifically, the \gls{m2} algorithm finds the affected walks (Line~\ref{alg:l:m2-filter}) and trims them to the first affected vertex (Line~\ref{alg:l:m2-trim}). After that, \gls{m2} resumes the trimmed random walks until they are of the given length $l$. 
% At Line~\ref{alg:l:m2-trim}, \gls{m2} trims the affected walks to the first occurrence of an affected vertex in $\vertices_e$.
% For the new affected vertices $V_n$, $\gls{m2}$ initializes random walks the same as in algorithm $\gls{m1}$ and then, merges the two sets of $W_e$ and $W_n$ (Lines~\ref{alg:l:m2-init} and \ref{alg:l:m2-merge}). The result set of random walks are given to the random walk function at Line~\ref{alg:l:m1-rw} that completes the random walk steps for all the given random walks up to length $l$. Finally, it updates the old random walks $W^{t}$ given new set of walks $W^{t+1}$, which adds the new walks and replaces the old walks for the existing walks.

Due to the computational expense of searching for the affected vertices in the affected walks, we also introduce the \gls{m3} algorithm that is the same as the \gls{m2} algorithm but instead of trimming the affected walks it re-generates the affected walks from their first step. However, as we show in the rest of this section, re-generating the affected walks from their first vertices create walks that are biased toward the walks that do not visit the affected vertices.

\begin{figure}[t]
\begin{center}
\subfigure{\includegraphics[width=\columnwidth]{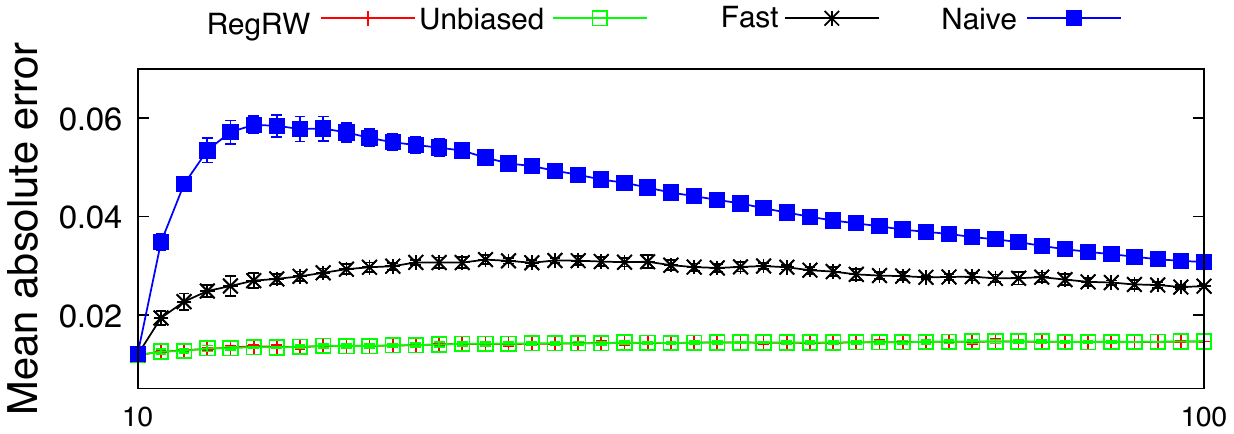}\label{fig:cora1-mean-error-m1-4}}\vspace{-1.5em}
\subfigure{\includegraphics[width=\columnwidth]{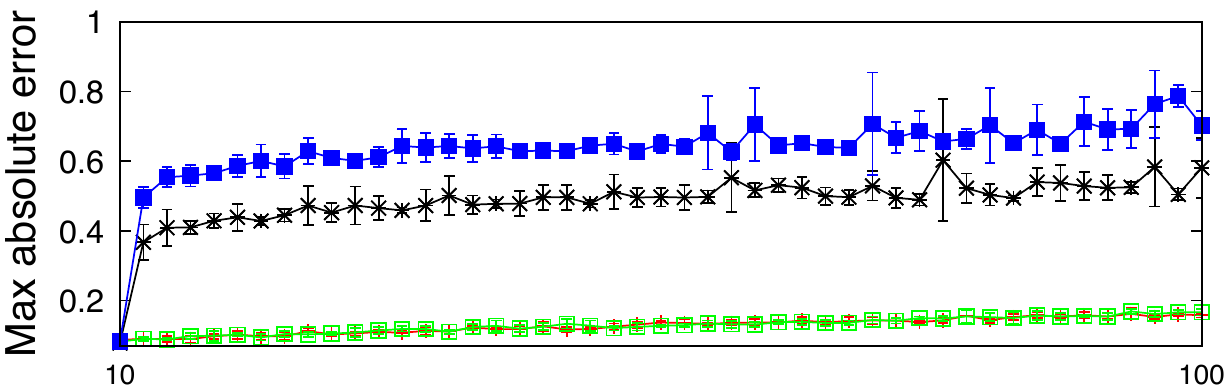}\label{fig:cora-max-error-m1-4}}\vspace{-1.5em}
\subfigure{\includegraphics[width=\columnwidth]{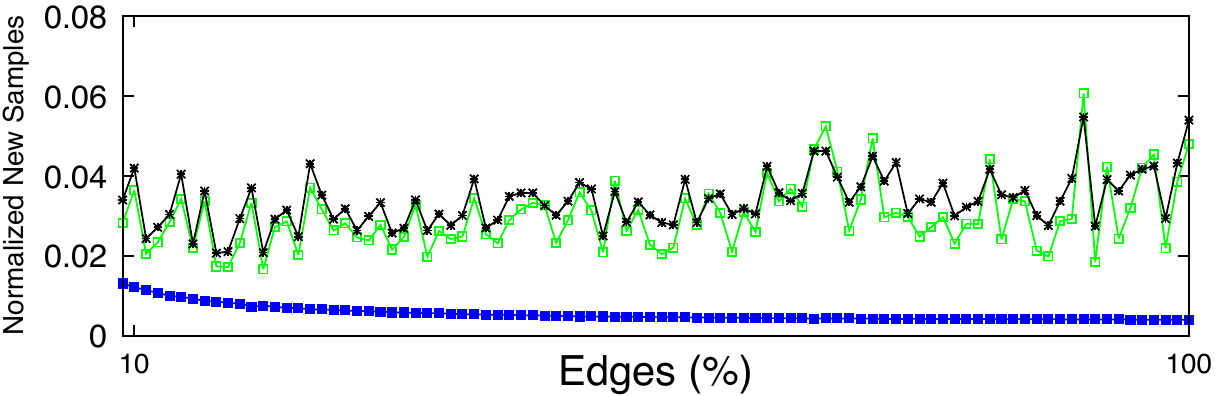}\label{fig:cora-m1-4-steps5}}
\end{center}
\caption{The mean and maximum absolute errors and the proportion of random walks that are updated for the proposed algorithms compared to the unbiased \gls{m1} algorithm. The experiment is done on the Cora dataset. The error represents the bias and is computed by subtracting the empirical estimates of the transition probability from the theoretical transition probability for each vertex. The figures show that the \gls{m3} algorithm is statistically indistinguishable from the \gls{m1} algorithm while requiring regeneration of less than $4\%$ of the random walks for each snapshot. We have observed a similar pattern on our other datasets.}
\label{fig:mean-errors}
\end{figure}

We have claimed that the \gls{m4} and \gls{m3} algorithms produce random walks that have biased statistics compared to fully updating the random walks. We now show this empirically 
by plotting the difference between the empirical transition probabilities of the pairs generated by the four random walk update algorithms calculated using \cref{eq:empirical_trans_probs} and the theoretical transition probabilities on the graph $G^{t+1}$.
These differences are shown in \cref{fig:mean-errors} which depicts the error in the transition probabilities and the normalised number of random walks (re)-generated by each algorithm compared to the number of random walks in the complete corpus $W_{G^{t+1}}$.
The errors are shown in terms of mean and maximum absolute errors for all the vertices when generating random walks on the Cora dataset. The error bars are the standard deviation over 5 runs of each experiment. At each run, the graph is initialized with $10\%$ of the edges and $5$ edges are added to the graph at each time step to build the next snapshot. The details about the datasets and the experiment setup are explained in \cref{sec:eval}. For each snapshot of the graph, we run the four algorithms and compute the errors. We can see that the \gls{m1} and \gls{m2} algorithms have a very small error compared to the \gls{m3} and \gls{m4} algorithms which are seen to give biased transition probabilities. We note that \gls{m4} has the highest error because it only generates random walks for new and affected vertices and does not consider the affected walks. 
We also note that the \gls{m2} algorithm can update the random walk corpus by updating less than $4\%$ of the random walk corpus compared to \gls{m1}.
%This means that \gls{m2} generates random walks that are statistically indistinguishable from \gls{m1} that re-generates random walks for all the vertices for each snapshot of the graph. 
In the rest of the experiments, we ignore \gls{m3} as it has no clear advantage over \gls{m2} except that it does not require searching the random walks for the first affected vertex. However, we note that by storing the random seed that is used to generate each random walk we can regenerate the random walks up to the first affected vertex by simply reusing this seed when the walk is regenerated. Hence, by storing the random seeds for each walk we can implement the \gls{m2} algorithm with the same computational cost as the \gls{m3} algorithm.

% As it can be seen, as the graph gets denser and more edges are added, the mean error in \gls{m4} is reduced. This is because the random walks can reach more vertices in the graph in a denser graph.
% Comparing \Cref{fig:cora1-mean-error-m1-4,fig:wiki-mean-error-m1-4} also confirms the effect of the density of graphs on the errors of the random walks. Since the Wikipedia dataset is 4 times denser than the Cora dataset, the error of the random walks created by the $M_4$ algorithm is closer to the $M_1$ and $M_2$ on the Wikipedia dataset. 
% This can be because of the number of Strongly Connected Components that decreases as more edges are added to the graph (?). 
% $\gls{m4}$ is similar to how Hamilton et. al~\cite{hamilton2017inductive} proposed to adapt DeepWalk for dynamic graphs with the difference that $\gls{m4}$ considers affected vertices in addition to the newly added vertices to the graph. However, $\gls{m3}$ and
% $\gls{m4}$ may generate biased walks and therefore not representing transition probabilities between nodes of the graph as expected by the random walk configuration parameters (see \cref{so-to-fo}). In \crefrange{alg:m1}{alg:m4}, the algorithms are presented.
% \begin{figure}[t]
% \begin{center}
% \subfigure[\scriptsize Mean absolute error.]{\includegraphics[width=0.5\textwidth]{pics/evals/cora-m1-4-errors-s5}\label{fig:cora1-mean-error-m1-4}}
% \subfigure[\scriptsize Max absolute error steps.]{\includegraphics[width=0.5\textwidth]{pics/evals/cora-m1-4-max-errors-s5}\label{fig:cora-max-error-m1-4}}
% \subfigure[\scriptsize New samples generated compared to \gls{m1}.]{\includegraphics[width=0.5\textwidth]{pics/evals/cora-m1-5-rw-steps5}\label{fig:cora-m1-4-steps5}}
% \end{center}
% \caption{The mean and max absolute errors and the number of new samples generated in different random walk algorithms compared to \gls{RegRW}.}
% \vspace{-1.5em}
% \label{fig:mean-errors}
% \end{figure}

\subsubsection{Complexity Analysis}
Consider a snapshot of graph $\graph^{t}=\{\vertices^{t},\edges^{t}\}$ at time $t$ with $n=|\vertices^{t}|$ and $m=|\edges^{t}|$ and a random walk corpus $W_{G^t}$ consisting of $r$ random walks starting at each vertex of the graph and having length $l$.
% We define the time complexity by number of times that the random walk algorithm computes random steps. 
The computation cost of the \gls{m1} algorithm to generate random walks for all the vertices in $\graph^{t}$ is $n \times r \times l$ steps of the random walk algorithm.

% It also needs to update the walk storage, which requires to write $n.r.l$ vertices. 
In \gls{m2}, the major functions are \emph{filter}, \emph{randomwalk}, and \emph{update} (see \cref{alg:m2}). For simplicity, we ignore the time complexity of the \emph{trim} function and instead we assume 
% it can be merged with filter function and also in the time complexity analysis
all walks have the same length of $l$ for the \emph{randomwalk} function. The computation complexity of \emph{filter} using a naive linear search is $O(a \cdot n \cdot l \cdot r)$, where $a=|\vertices^{1}_{\affected}|$ is the number of affected vertices. However, \emph{filter} can be implemented faster by using advanced search techniques such as reverse-indexes on vertices. Note that the \emph{filter} function is embarrassingly parallel and it can be done efficiently on cluster-computing frameworks such as Apache Spark~\cite{zaharia2010spark}. The time complexity of \emph{randomwalk} is $O(W \cdot l)$, where $W$ is the number of affected and new walks and $r\le W \le n \cdot r$. It is important to note that $l,r \ll n$. Usually, $l$ and $r$ are in the order of tens while $n$ is in the order of millions to billions, which is several orders of magnitude larger than $l$ and $r$. The smallest update on a graph is when a vertex is updated which incurs $r$ number of walks with length $l$. However, the upper bound $n \cdot l \cdot r$ in \gls{m2}, is a trivial upper bound that occurs when all the walks on the graph $G^t$ are affected, namely all vertices have an edge that has been either added or deleted. 
For \gls{m4}, the computation cost depends only on the affected vertices, i.e., $O(a\cdot l \cdot r)$.

\subsection{Updating Vertex Representations}

% [AD] This table confused me totally. Are you just saying that all combinations are possible? Does that need a table??
% \begin{table}[t]
% \caption{Possible combinations of the algorithms in order to generate random walk samples and to train the skip-gram model for node representation learning in a dynamic graph. Deepwalk and are based on re-generating all random walks and training the skip-gram model from scratch, i.e., a combination of \gls{m1} and \gls{u1}.}
% \scalebox{0.7}{
% \begin{tabular}{ |c|c|c|c|c| }
%  \hline
%  \backslashbox{Skip gram}{Random walk} & \gls{m1} & \gls{m2} & \gls{m3} & \gls{m4}\\
%  \hline
%  \gls{u1} & DeepWalk/node2vec & \checkmark & \checkmark & \checkmark\\
%  \hline
%  \gls{u2} & - & \checkmark & \checkmark & \checkmark\\
%  \hline
% \end{tabular}}
% \vspace{-1.5em}
% \label{table:algorithms-combinations}
% \end{table}

Now we consider the problem of updating the vertex representations to give new representations that capture the current state of the graph. Specifically, we define the problem as updating the vertex representations given an updated corpus of random walks $W_{G^{t+1}}$ and the previous representations of the vertices in the graph $z^{t}_k$ for $k\in \{0, \ldots, n(t)\}$ that were generated on the previous graph snapshot $G^t$.

The baseline algorithm to learn vertex representations is to optimize the objective function \cref{eq:unsup-obj} using the skip-gram model of~\cite{mikolov2013efficient}. The algorithm proceeds by initialising the vertex representations randomly. Then, the algorithm performs stochastic gradient descent to optimise the function \cref{eq:unsup-obj} or its negative sampling version over the training data consisting of the target-context pairs in the corpus $C_p(W_{G^{t+1}})$ created by the updated random walks $W_{G^{t+1}}$. For simplicity, we call this baseline algorithm \gls{u1}.

However, \gls{u1} is computationally expensive and inefficient for dynamic graphs, as it needs to train the skip-gram model from a random initialisation using the entire corpus $C_p(W_{G^{t+1}})$ regardless of the extent of the change in the graph. In \cref{subsec:dyanamic-rw}, we see that the number of random walks that algorithms \gls{m2}, \gls{m3}, and \gls{m4} create depends on the extent of the change on the graph.

Therefore, we propose to use this information in training the skip-gram model as well.
The \gls{u2} algorithm starts with the vertex representations from the previous graph snapshot $z^{t}_k$ and introduces randomly initialised vectors for the new vertices in the snapshot $A_{\vertices^{t+1}}$. Stochastic gradient descent is then performed to minimize the skip-gram objective function using the newly generated random walks for graph $\graph^{t+1}$.

Intuitively, as the optimisation process has not seen the regenerated affected walks, using the vertex pairs generated from the updated affected walks to minimise the skip-gram objective will allow updating the vertex representations with the changes that have occurred in the graph. This is, in essence, a transfer learning process~\cite{bengio2012deep} for the graph where the previously learned representations are fine-tuned using the vertex pairs from the updated walks, with a suitable specified learning rate.
% \textcolor{blue}{Cite: Y Bengio "Deep learning of representations for unsupervised and transfer learning" Workshop on Unsupervised and Transfer Learning, 2012 - jmlr.org}
In addition, this is highly related to few-shot learning which seeks to update a learning objective that has been previously trained on a large corpus of data with a small number of examples of a new class~\cite{ravi2016optimization}. We reserve as future work the investigation about an approach similar to the few-shot learning approach of ~\cite{ravi2016optimization} for updating the vertex representations.
% could be training a meta-learner to update the vertex representations given the new ones similar to the few-shot learning approach of

% \textcolor{red}{Note: in the future a more principled approach could be training a meta-learner to update the vertex representations given the new ones similar to the few-shot learning approach of Ravi \& LaRochelle: https://openreview.net/pdf?id=rJY0-Kcll}

% However, only providing the learning algorithm the updated affected random walks means that the learning algorithm will see a biased representation of the graph statistics which could lead to the updated representations performing poorly in the downstream task.
% To explore this we also propose an algorithm which samples some of the unaffected walks as well. We define the parameter $O$ that represents the number of unaffected random walks that are randomly sampled as a proportion of the number of affected walks.

As we show in \cref{sec:eval}, the \gls{u2} algorithm can learn vertex representations competitive to \gls{u1} with much less training samples.

%% file: chapters/algorithms/m1.tex
\begin{algorithm}[t]
\caption{\gls{m1}}
\footnotesize
\label{alg:m1}
\begin{algorithmic}
\Procedure{\gls{m1}}{$\graph^{t+1}$, $r$, $l$}
    \State $V \leftarrow \text{vertices}(\graph^{t+1})$
    \State $W \leftarrow \text{initWalks}(V,r)$ \label{alg:l:m1-init}
    \State $W^{t+1} \leftarrow \text{randomwalk}(W,l,\graph^{t+1})$ \label{alg:l:m1-rw}
    \State \Return $W^{t+1}$
\EndProcedure
\end{algorithmic}
\end{algorithm}

%% file: chapters/algorithms/m4.tex
\begin{algorithm}[t]
\caption{\gls{m4}}
\footnotesize
\label{alg:m4}
\begin{algorithmic}[1]
\Procedure{\gls{m4}}{$\graph^{t+1}, W^{t}, \vertices^{1}_{\affected}, r, l$}
    \State $W \leftarrow \text{initWalks}(\vertices^{1}_{\affected},r)$
    \State $W^{t+1} \leftarrow \text{randomwalk}(W,l,\graph^{t+1})$
    \State $W^{t+1} \leftarrow \text{update}(W^{t}, W^{t+1})$ \label{alg:l:m4-update}
    \State \Return $W^{t+1}$
\EndProcedure
\end{algorithmic}
\end{algorithm}

%% file: chapters/algorithms/m2.tex
\begin{algorithm}[t]
\caption{\gls{m2}}
\footnotesize
\label{alg:m2}
\begin{algorithmic}[1]
\Procedure{\gls{m2}}{$\graph^{t+1}, W^{t}, \vertices^{1}_{\affected}, r, l$}
    \State $\vertices_n \leftarrow \text{newVertices}(\vertices^{1}_{\affected})$
    \State $\vertices_e \leftarrow \text{existingVertices}(\vertices^{1}_{\affected})$
    \State $W_{affected} \leftarrow \text{filter}(W^{t}, \vertices_e)$ \label{alg:l:m2-filter}
    \State $W_e \leftarrow \text{trim}(W_{affected}, \vertices_e)$ \label{alg:l:m2-trim}
    \State $W_n \leftarrow \text{initWalks}(\vertices_n,r)$ \label{alg:l:m2-init}
    \State $W \leftarrow W_e \cup W_n$ \label{alg:l:m2-merge}
    \State $W^{t+1} \leftarrow \text{randomwalk}(W,l,\graph^{t+1})$ \label{alg:l:m2-rw}
    \State $W^{t+1} \leftarrow \text{update}(W^{t}, W^{t+1})$ \label{alg:l:m2-update}
    \State \Return $W^{t+1}$
\EndProcedure
\end{algorithmic}
\end{algorithm}

%% file: chapters/06-evaluation.tex
\section{Experiments}\label{sec:eval}
We have implemented the random walk algorithms and the target-context pairs generator in Scala, and the skip-gram model in python with TensorFlow framework~\cite{abadi2016tensorflow}\footnote{https://github.com/shps/incremental-representation-learning}. We run the experiments a machine with 8 CPUs and 150GB memory.

The datasets used in the experiments are as follows:
\begin{itemize}
    \item Cora~\cite{wiki-dataset,mccallum2000automating}: is based on scientific papers citations. Each paper belongs only to one category among $7$ categories: Neural Networks, Reinforcement Learning, Probabilistic Methods, Genetic Algorithms, Rule Learning, Theory, and Case-Based Reasoning.
    \item Wikipedia~\cite{wiki-dataset}: is the dataset of the Wikipedia web pages from $17$ categories and the links between them.
    \item BlogCatalog~\cite{zafarani2009social}: is a network of bloggers and social relationships among them on the BlogCatalog website. The bloggers are labeled with at least one label that represents interests of the bloggers. The labels are extracted from the metadata provided by the bloggers.
    \item CoCit~\cite{ms2016,tsitsulin2018verse}: is a subgraph of the Microsoft Academic Graph, that is a network of academic papers. The dataset is a network of papers citing other papers, in which labels represent the conference that papers were published in.
\end{itemize}
% We remove the redundant edges \textcolor{green}{What are 'redundant edges'?} and the disconnected vertices, and make sure that the datasets consist of a single connected component. \textcolor{green}{Why do we ensure they consist of a single CC?} 
\cref{table:datasets} presents the statistics of vertices, edges, labels and the density of the datasets, which are represented as undirected graphs. 

\begin{table}[t]
\centering
\caption{Datasets: number of vertices $\lvert V \rvert$, number of edges $\lvert E \rvert$, Labels represent the number of categories in each dataset, Density is defined as $\lvert E \rvert/\binom{\lvert V \rvert}{2}$.}
\scalebox{0.9}{
\begin{tabular}{ ccccc } 
 Name & $\lvert V \rvert$ & $\lvert E \rvert$ & Labels & Density\\
 \hline
 Cora & 2,485 & 5,069 & 7 & $1.6\times10^{-3}$\\
 Wikipedia & 2,357 & 11,592 & 17 & $4.2\times10^{-3}$\\ 
 BlogCatalog & 10,312 & 333,983 & 39 & $6.3\times10^{-3}$\\
 CoCit & 42,452 & 194,410 & 15 & $2.0\times10^{-4}$\\
\end{tabular}}
\vspace{-1.5em}
\label{table:datasets}
\end{table}

The graphs do not contain temporal information, so we follow the previous research~\cite{du2018dynamic,li2017attributed} by creating an initial graph from a randomly selected subset of edges and at each step adding a specified number of randomly selected edges to the initial graph in order to create a new snapshot of the graph. We call the number of edges that are added to the graph at each step the \emph{update rate}. We run each experiment multiple times and report the mean of the results. The vertex representations are given to a one-vs-rest logistic regression classifier. The classifier is set to split the train and test data 10 times and we present the mean Macro-F1 score. The standard deviations for the accuracy are less than $0.01$ unless it is explicitly mentioned. 

Throughout the experiments, the walk lengths and the skip-gram window size are set to $10$ and $8$ respectively. The number of walks and the size of vector representation for each vertex are set to $80$ and $128$ respectively as in previous research~\cite{perozzi2014deepwalk}. 
% \textcolor{green}{I think in deepwalk the walk length was 80 not the number of walks. Note that we have shown there is little difference between them} 
The mini-batch size for the skip-gram model training is set to $200$. The initial learning rate used is to $0.025$ for the Cora, Wikipedia, and CoCit datasets. For the BlogCatalog dataset, we observed a better performance when the learning rate is set to $0.2$. We set the number of epochs to $3$ for the Cora and Wikipedia datasets, but $10$ and $4$ for the BlogCatalog and the CoCit datasets respectively. 
% \textcolor{green}{We may need to justify this as 3 epochs is not very many. Is this for incremental training or for training from scratch for the static algorithm? How do the learning rates change with static training vs incremental training?} 

% For the BlogCatalog dataset, we see the results can be slightly improved by increasing the number of epochs to more than $10$. However, since the number of epochs affects the performance of the algorithms in the same way we do not increase the number of epochs to more than 10. \textcolor{green}{Why is this true only for BlogCatalog? The performance doesn't increase with Cora for more than 3 epochs?}

\subsection{Vertex Classification}
\begin{figure*}[t]
\begin{center}
\subfigure[\scriptsize Cora]{\includegraphics[width=0.3\textwidth]{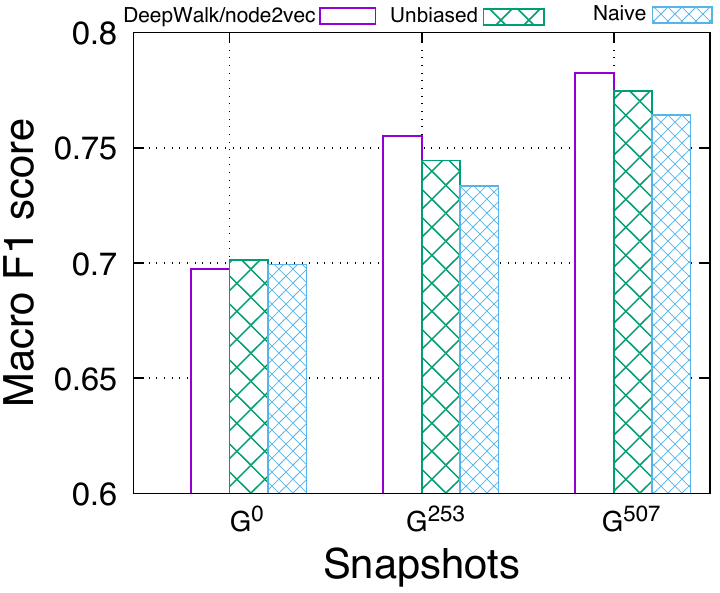}\label{fig:cora-f1-snapshots}}
\subfigure[\scriptsize Wikipedia]{\includegraphics[width=0.3\textwidth]{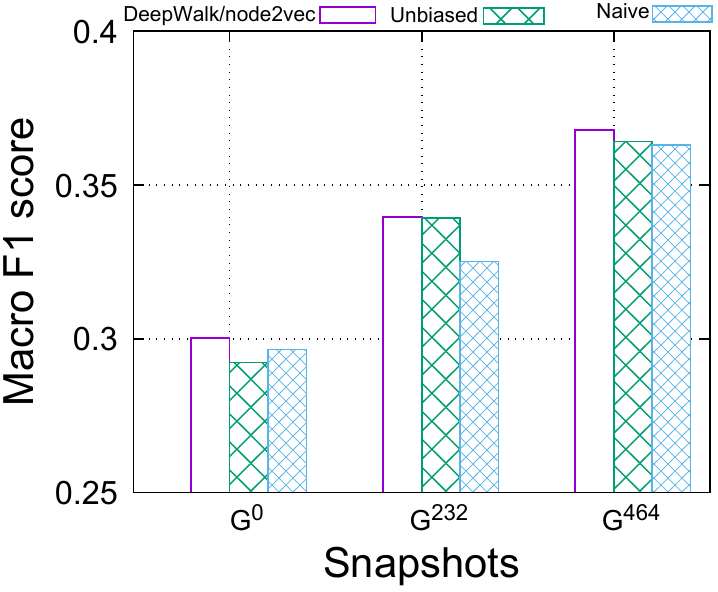}\label{fig:wiki-f1-snapshots}}
\subfigure[\scriptsize Cocit]{\includegraphics[width=0.3\textwidth]{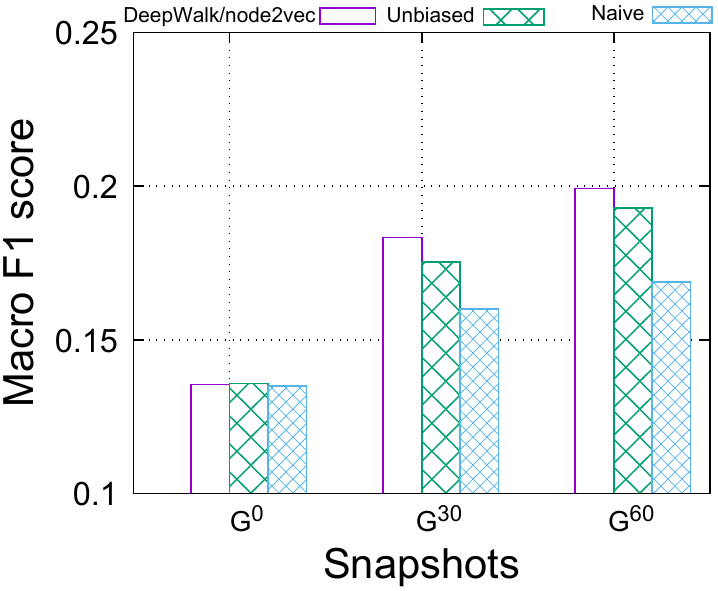}\label{fig:cocit-f1-snapshots}}
\end{center}
\caption{Multi-class vertex classification results for different embedding methods on different datasets. The vertex representations of random walk algorithms \gls{m2} and \gls{m4} are learned by \gls{u2} method with parameter $O=100\%$. The DeepWalk is equivalent to learn embeddings by using \gls{m1} and \gls{u1}.  We show the average results for the first, middle, and the last snapshot in all runs. We initialise the first snapshot $G^0$ for each dataset with a different number of edges with respect to their total number of vertices. The initial number of edges for Cora (a), Wikipedia (b), and CoCit (c) are $50\%$, $10\%$, and $4\%$ accordingly. Based on the number of edges, we set the update rate to 5 for the Cora and Wikipedia datasets and 200 for the CoCit dataset.}
\vspace{-1.5em}
\label{fig:f1-snapshots}
\end{figure*}

\begin{figure*}[t]
\begin{center}
\subfigure[\scriptsize Cora]{\includegraphics[width=0.3\textwidth]{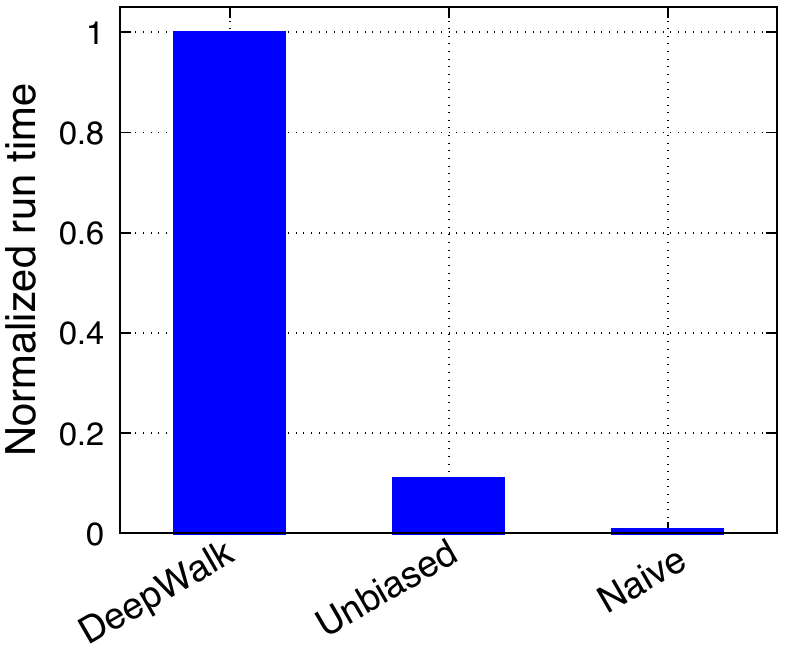}\label{fig:cora-time-snapshots}}
\subfigure[\scriptsize Wikipedia]{\includegraphics[width=0.3\textwidth]{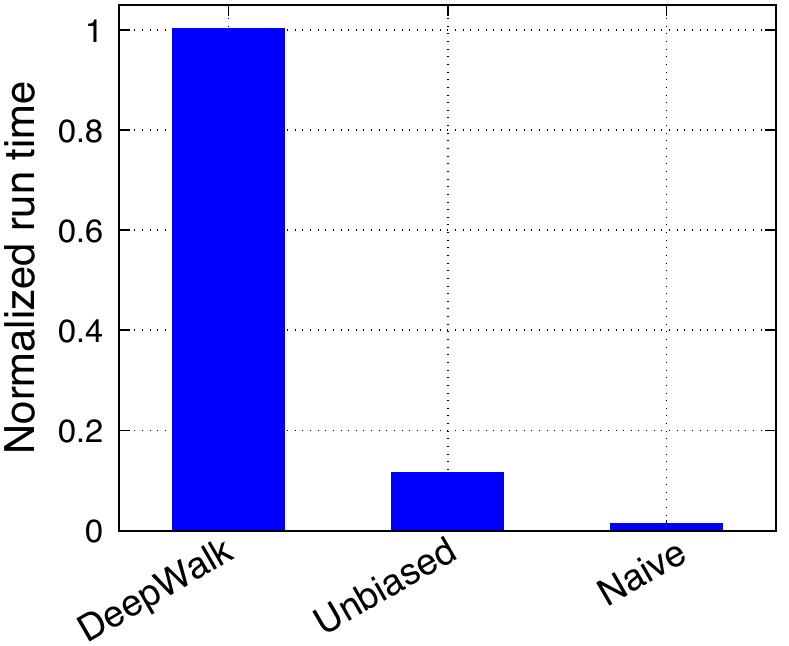}\label{fig:wiki-time-snapshots}}
\subfigure[\scriptsize Cocit]{\includegraphics[width=0.3\textwidth]{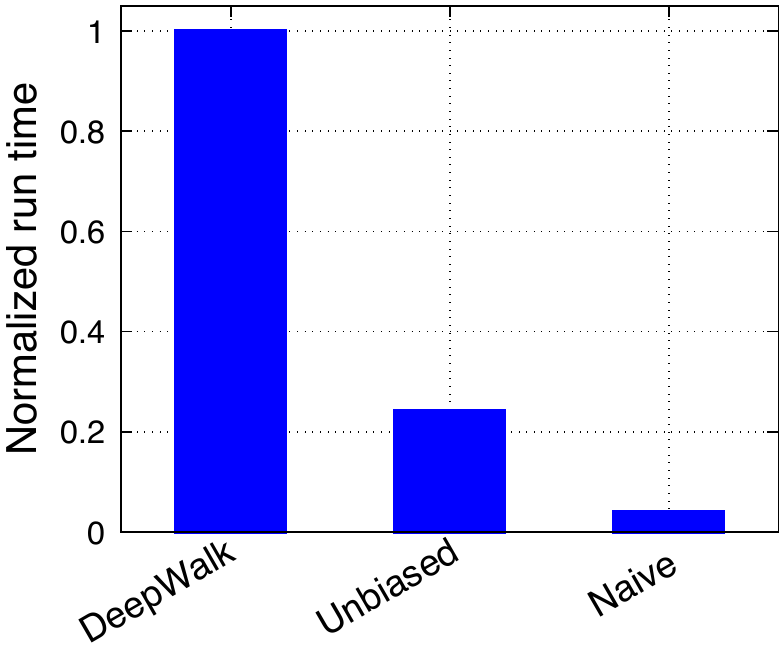}\label{fig:cocit-time-snapshots}}
\end{center}
\caption{Run time of different methods normalized to the run time of DeepWalk (\gls{m1} and \gls{u1}). As can be seen, our incremental methods are several times faster than DeepWalk. The run times belong to the last snapshot of the graphs in \cref{fig:f1-snapshots}. The run time includes the time for generating random walks and training the skip-gram model. We use \gls{u2} to learn embeddings for our algorithms \gls{m2} and \gls{m4}.}
\vspace{-1.5em}
\label{fig:time-snapshots}
\end{figure*}

We evaluate the performance of the proposed dynamic graph algorithms for learning vertex representations, i.e., the \gls{m2} and \gls{m4} algorithms using the \gls{u2} method to update the skip-gram model. We compare these dynamic representation learning algorithms against our implementation of DeepWalk~\cite{perozzi2014deepwalk}/node2vec~\cite{grover2016node2vec}, which re-generates all the random walks and re-trains the skip-gram model. Specifically, DeepWalk and node2vec are equivalent to the \gls{m1} algorithm combined with the \gls{u1} method.

Network snapshots in a dynamic graph often contain multiple disconnected components. Typically, for many real-world social network datasets such as the DBLP co-authorship network~\cite{dblp-ds2018} the snapshots consist of a connected component with the majority of vertices and many smaller disconnected components.
We have also observed the same behaviour when adding edges in random batches for all the datasets used in this paper.
As the node representations learned on disconnected components do not have any relation to each other any downstream task must be performed separately on each disconnected component~\cite{hamilton2017representation}. Therefore, while at every step we learn embeddings for all the components in the current snapshot, we train and evaluate the classifier only for the nodes of the largest connected component.

We evaluate the performance of the downstream classification task by using cross-validation on $9\%$ labelled data and $10$ times re-shuffling and splitting data into train/test sets. Evaluating the performance of the downstream classifier requires a considerable number of vertices for train/test splits. Therefore, we create the initial snapshots of the graphs by randomly selecting a proportion of edges. To that end, we initialised the Cora, Wikipedia, and CoCit graphs with $50\%$, $10\%$, and $4\%$ of the edges and created consecutive snapshots of the graphs by adding more edges.

The results for the different methods are shown in \Cref{fig:f1-snapshots} which shows the performance of the downstream classification task as measured by the cross-validated Macro-F1 score.
The performance scores are shown for the initial graph snapshot and for two snapshots chosen from the middle and the end of the experiment. 
% The final snapshot is equivalent to the entire graph given by each dataset which is connected due to the pre-processing performed on each graph.
% [AD] What trend are you talking about? I don;t think we need that sentence.
%We have observed the same trend for the results of the other snapshots in between of the presented snapshots.
We see that both dynamic update algorithms give representations that have competitive performance as compared to the full \gls{m1} algorithm for each snapshot of the graph.
In addition, we see that the \gls{m2} algorithm gives marginally better performance in the downstream task than the \gls{m4} algorithm. This is consistent with the fact that the \gls{m4} algorithm produces random walks that are biased in terms of the pair-wise transition probabilities, as discussed in \cref{subsec:dyanamic-rw-problem}.

\cref{fig:time-snapshots} shows the total run time for each of the random walk and skip-gram training algorithms on the final snapshot of the graphs normalised to the run time of DeepWalk. The dynamic methods are, as expected, $9$ times up to $160$ times faster than DeepWalk (\cref{fig:time-snapshots}). 

\subsection{Update Rate}
As we explain in \cref{subsec:dyanamic-rw}, the computational complexity of the \gls{m4} random-walk algorithm is proportional to the number of affected vertices in the graph update. This is compared to the \gls{m2} algorithm where the computational complexity is proportional to the number of affected walks. As the number of affected walks will be greater than the number of affected vertices the \gls{m4} algorithm is computationally more efficient than the other algorithms (as can be seen in \cref{fig:time-snapshots}).

\begin{figure}[t]
\centering
\includegraphics[scale=0.8]{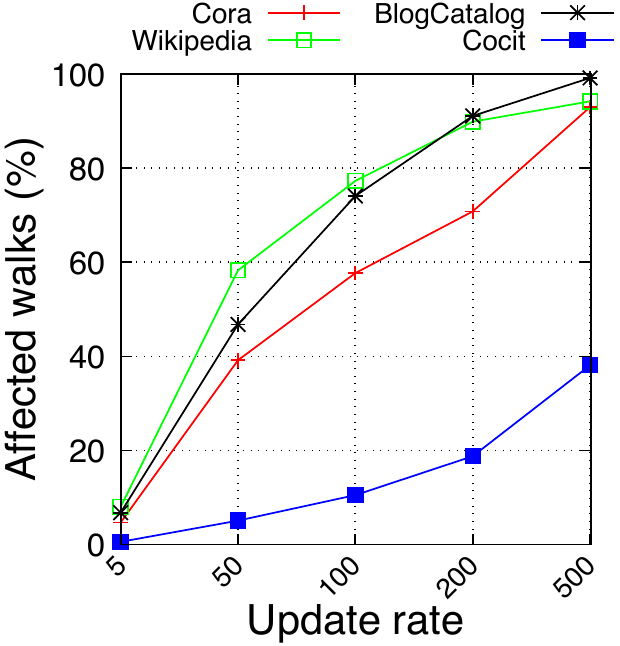}
% \vspace{-0.5cm}
\caption{The percentage of affected walks as the update rate increases on different datasets. The graphs are initialised with $90\%$ of the total number of edges. It can be seen that the percentage of affected walks depends on the density of the graphs. The highest density dataset, BlogCatalog, shows the highest growth in the number of random walks being affected for each update of the graph as the update rate increases. For datasets with lower density, such as CoCit, the growth in the number of random walks that are affected at each graph update.}
\label{fig:affected-datasets}
\end{figure}

The run time of the \gls{m2} algorithm depends on the number of affected walks which in turn depends on the density of the graphs as well as the number of affected vertices. In our experiments, the number of affected vertices is approximately proportional to the update rate.
To illustrate this for the datasets used in this paper, \cref{fig:affected-datasets} shows the percentage of affected walks as a proportion of all the walks generated as the update rate increases. The graphs are initialised with $90\%$ of the edges. As expected, increasing the update rate increases the number of affected walks and the denser a graph is the larger the number of affected walks for a given update rate. For example, the CoCit dataset has the lowest density of the datasets and hence we see that the proportion of affected walks is smaller for the same update rate as compared to the other datasets.

The effect of this is for a given update rate, \gls{m2} performs better for the graphs with low density. Therefore, dynamic large social graphs are naturally good targets to use our algorithm \gls{m2} as they usually have at least one order of magnitude lower density than the CoCit graph~\cite{network-storage18}.

\begin{figure}[t]
\begin{center}
% \subfigure[\scriptsize \gls{m2} on Cocit]{\includegraphics[width=0.13\textwidth]{pics/evals/cocit-affecteds-m2}\label{fig:cocit-affected-m2}}
\subfigure[\scriptsize \gls{m2}]{\includegraphics[width=0.23\textwidth]{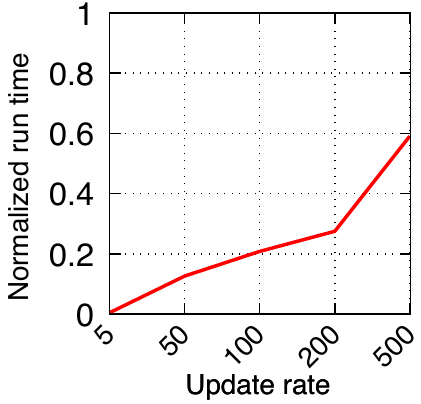}\label{fig:cocit-rw-time}}
\subfigure[\scriptsize \gls{u2}]{\includegraphics[width=0.23\textwidth]{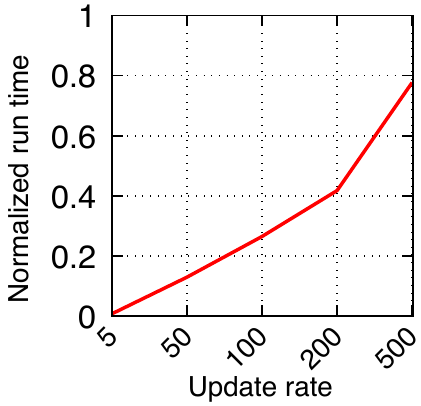}\label{fig:cocit-w2v-time}}
\end{center}
\caption{The run time of \gls{m2} and \gls{u2} normalized to the run time of \gls{m1} and \gls{u1} as the update rate increases on the CoCit dataset. The run times belongs to the final snapshot of the graph with $100\%$ of edges. The figure shows that the run time of our incremental algorithm increases as the update rate increases. This is because by increasing update rates a higher portion of graph elements will be affected. Therefore, our algorithms \gls{m2} and \gls{u2} need to re-generate a higher number of samples that will affect the run time.}
\vspace{-1.5em}
\label{fig:cocit-run-time}
\end{figure}

\cref{fig:cocit-run-time} depicts the run time of the \gls{m2} $+$ \gls{u2} method normalised to the run time of the \gls{m1} $+$ \gls{u1} method. The results are for the last snapshot of the CoCit graph where the entire dataset has been streamed to the algorithms. We can see that when update rate is $5$ edges per snapshot, the \gls{m2} $+$ \gls{u2} method is two orders of magnitude faster than the \gls{m1} and \gls{u1} method. 
% Note that the results are for $U_2$ when sub-sampling rate $O$ is set to $100$. The computation time for $U_2$ is almost halved when sub-sampling rate $O$ is set to $0$.
% [AD] This is clear as you have just doubled the number of vertex pairs fed to the algorithm. I had intended you to *fix* the number of vertex pairs and have $O$ be the proportion of that fixed number of vertex pairs that is sampled from the unaffected nodes. Thus at $0=0$ you would have all affected nodes, at $O=50$ you would have 50/50 split of unaffected and affected nodes BUT THE SAME TOTAL NUMBER. Otherwise, we aren't comparing apples to apples.

\subsection{Random Walks and Bias in the \gls{m4} Algorithm}
\begin{figure*}[t]
\begin{center}
\subfigure[\scriptsize Cora]{\includegraphics[width=0.23\textwidth]{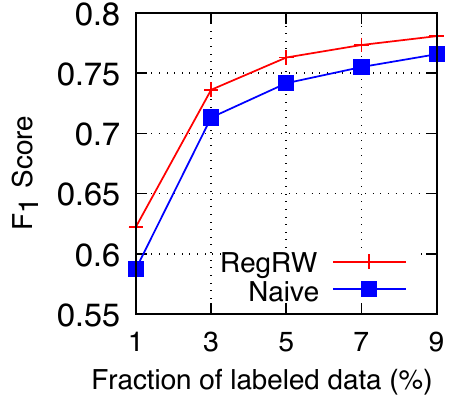}\label{fig:cora-f1-labeled}}
\subfigure[\scriptsize Cocit]{\includegraphics[width=0.23\textwidth]{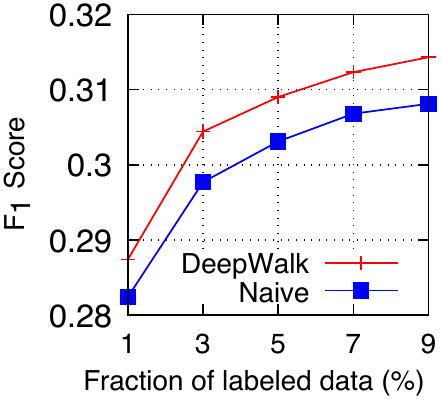}\label{fig:cocit-f1-labeled}}
\subfigure[\scriptsize Wikipedia]{\includegraphics[width=0.23\textwidth]{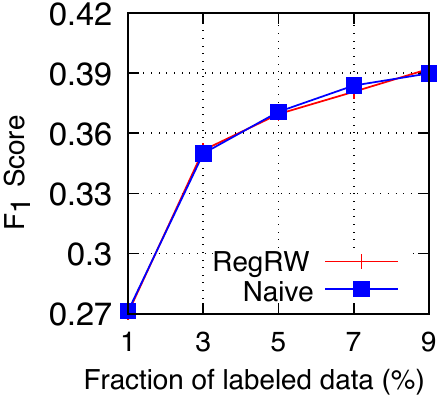}\label{fig:wiki-f1-labeled}}
\subfigure[\scriptsize BlogCatalog]{\includegraphics[width=0.23\textwidth]{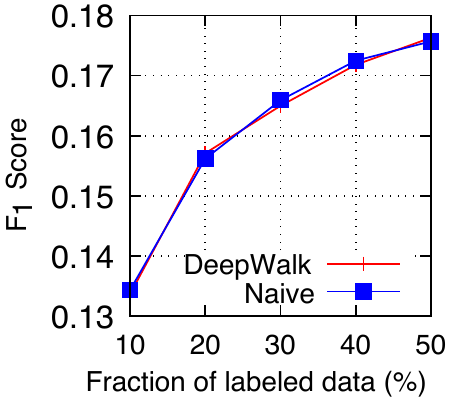}\label{fig:blog-f1-labeled}}
\end{center}
\caption{The effect of bias in the random-walk corpus generated by the \gls{m4} algorithm on multi-class classification (for datasets Cora, Wikipedia, and CoCit) and multi-label classification (for dataset BlogCatalog). The results are shown in terms of Macro F1-score. The results are shown for different percentage of labelled data for training the classifier. The graphs are initialised with $10\%$ of edges and the update rate is 5. We run our algorithm \gls{m4} for each snapshot of the graphs until the last snapshot, where it has $100\%$ of edges. The vertex representations are learned by the \gls{u1} method with the random walk samples generated by the \gls{m4} and \gls{m1} algorithms on the final snapshot of each graph. Note that we run the same experiment for the \gls{m2} algorithm. As both the \gls{m1} and \gls{m2} algorithms produce a corpus of unbiased random walks, we only present the results for the \gls{m1} algorithm. As the results show, the \gls{m4} algorithm performs as good as the \gls{m1} algorithm for the Wikipedia and BlogCatalog datasets. However, the performance of the \gls{m4} algorithm is slightly inferior to the \gls{m1} algorithm for Cora and CoCit. Furthermore, in all the experiments the results are stable as we change the amount of training data.}
\vspace{-1.5em}
\label{fig:m1-4-f1-labeled}
\end{figure*}

As discussed in \cref{subsec:dyanamic-rw}, \gls{m4} generates random walks that are biased in terms of the pair-wise transition probabilities.
To evaluate the effect of the biased random walks produced by the \gls{m4} algorithm, we re-train the skip-gram model starting with a random initialisation of the weights and use all the vertex pairs generated from the random walk algorithms in the last snapshot of each graph, where each graph has $100\%$ of the edges. 
% We also add the results for a \emph{random} classifier that predicts only based on the distribution of labels in the training data. 
We run each experiment $10$ times and the mean Macro F1-score of the results for multi-class and multi-label classifications are presented in \cref{fig:m1-4-f1-labeled}. The performance scores for the \gls{m2} algorithm are similar those for the \gls{m1} algorithm. The random-walk corpus generated by \gls{m4} is biased and this seems to give a negative effect on the downstream task performance for Cora and CoCit data sets (\cref{fig:cora-f1-labeled} and \cref{fig:cocit-f1-labeled}) as seen in  \cref{fig:m1-4-f1-labeled} (a) and (b). However, we don't observe such negative effect of the random-walks generated by the \gls{m4} algorithm on the Wikipedia and BlogCatalog datasets (\cref{fig:wiki-f1-labeled} and \cref{fig:blog-f1-labeled}).
%  Furthermore, when the accuracy of the results are lower the effect of biased walks are less evident. The same characteristics holds for the BlogCatalog dataset and the CoCit datasets (see \cref{table:bc-cc-f1}).
This could be due to fact that the number of random walks that are updated by the \gls{m4} algorithm is higher for graphs with higher density. In addition, the fact that the random walk corpus used to update the vertex representations is statistically biased does not necessarily affect the downstream task accuracy if this bias is small and localised. In the future, it would be of interest to investigate the differences in the representations that are found by the biased versus unbiased methods for different datasets.

%% file: chapters/07-related-work.tex
\section{Related Work}
Feature engineering has been long studied for graph analysis tasks. It requires domain experts to extract features~\cite{henderson2011s} for vertices by handcrafting features and using feature extraction techniques~\cite{tang2012unsupervised}. In contrast, the focus of this paper is on the general-purpose representation learning approaches.
% \textcolor{blue}{I'm not sure we need to say this.}

Representations learning for graphs is an important problem due to its application for tasks such as link prediction~\cite{backstrom2011supervised} and vertex classification~\cite{tsoumakas2007multi}. Many of the initially proposed techniques~\cite{belkin2002laplacian,tenenbaum2000global,roweis2000nonlinear} are to learn graph representations based on a spectral analysis of the adjacency matrix. 

More recently the first and second order proximity of vertices have been used to learn vertex representations by \cite{tang2015line} by modelling the joint probabilities and the transition probabilities of connected vertices.
Driven by recent advancements in word embedding~\cite{mikolov2013efficient,mikolov2013distributed} there have been a number of new methods to learn vertex representations in a similar way by using random walks in the graph as sentences in the same way as sentences in the word embedding methods~\cite{perozzi2014deepwalk,grover2016node2vec}.
In addition, there are methods~\cite{dong2017metapath2vec,chen2017task} that extend these methods to heterogeneous graphs.
A recent method VERSE~\cite{tsitsulin2018verse} proposes a more flexible approach for similarity notion than of local neighbourhood. VERSE explicitly learns any similarity measure among vertices, such as personalised PageRank, to learn vertex representations.

Another approach to learning vertex representations is GraphSAGE~\cite{hamilton2017inductive} which learns a neural-network that transforms the features of a vertex and a sampled sub-graph around that vertex to a vertex representation. Specifically, GraphSAGE takes an inductive approach that can be used to generate vertex representations for vertices that are not in the training graph. Other methods offer a similar approach that requires graphs that have features for each vertex~\cite{li2017attributed,li2018streaming}. 

% \textcolor{pink}{Graphsage is not about dynamic graphs. They may mention them, but so does the DeepWalk paper.} 

We note that all of the aforementioned methods are only for static graphs.
Du et al. extends LINE~\cite{tang2015line} to support dynamic graphs. Neighbourhood sampling methods based on random walks, that are the focus of this paper, have been shown to perform better than one-step and two-steps sampling in LINE~\cite{perozzi2014deepwalk,grover2016node2vec}. Chang et al.~\cite{chang2017streaming} propose a real-time recommender system on streaming data. In this paper, we focus on graph data and vertex representations that can be used for different downstream learning tasks.

Recently, a framework based on a modification of DeepWalk for temporal graphs has been proposed ~\cite{nguyen2018continuous} that uses temporal random walks where at each step the walker is only allowed to move along an edge that has a later time than the edge used to arrive at the current vertex.
The framework of ~\cite{nguyen2018continuous} aims to use temporal information after the fact, whereas in this paper we focus on updating vertex representations given new edges and vertices not previously available.
% This approach is different from the purpose of this paper as it aims to use temporal information after the fact whereas in this paper we focus on updating vertex representations given new edges and vertices not previously available.

% \textcolor{pink}{It would be good to refer to \& cite this recent review paper \cite{zhang2018review}}

%% file: chapters/08-conclusion.tex
\section{Conclusion}
Many of the real-world graphs are dynamic and change over time. However, the contemporary methods for unsupervised representation learning of vertices are mainly for static graphs. 
In this paper, we focused on efficient representation learning methods based on random walks for dynamic graphs. We discussed that naive incremental update of random walks results in random walks that statistically do not represent the graph structure when the graph changes over time. We proposed an intuitive way to capture changes in a dynamic graph based on the notions of affected vertices and affected walks. Following that, we proposed an incremental random walk algorithm, namely \gls{m2}, and an incremental method for representation learning, namely \gls{u2}, which their computation cost depends on the extent of the change in the graph. Incremental random walks generated by the \gls{m2} algorithm are statistically indistinguishable from re-generating random walks by the \gls{m1} algorithm. Through extensive experiments on real-world graphs, we showed that our incremental algorithms can achieve competitive results to the state-of-the-art methods while being considerably more efficient. 

%% file: chapters/09-acknowledgement.tex
% \section*{Acknowledgement}

%% file: ms.bbl
%%% -*-BibTeX-*-
%%% Do NOT edit. File created by BibTeX with style
%%% ACM-Reference-Format-Journals [18-Jan-2012].

\begin{thebibliography}{00}

%%% ====================================================================
%%% NOTE TO THE USER: you can override these defaults by providing
%%% customized versions of any of these macros before the \bibliography
%%% command.  Each of them MUST provide its own final punctuation,
%%% except for \shownote{}, \showDOI{}, and \showURL{}.  The latter two
%%% do not use final punctuation, in order to avoid confusing it with
%%% the Web address.
%%%
%%% To suppress output of a particular field, define its macro to expand
%%% to an empty string, or better, \unskip, like this:
%%%
%%% \newcommand{\showDOI}[1]{\unskip}   % LaTeX syntax
%%%
%%% \def \showDOI #1{\unskip}           % plain TeX syntax
%%%
%%% ====================================================================

\ifx \showCODEN    \undefined \def \showCODEN     #1{\unskip}     \fi
\ifx \showDOI      \undefined \def \showDOI       #1{#1}\fi
\ifx \showISBNx    \undefined \def \showISBNx     #1{\unskip}     \fi
\ifx \showISBNxiii \undefined \def \showISBNxiii  #1{\unskip}     \fi
\ifx \showISSN     \undefined \def \showISSN      #1{\unskip}     \fi
\ifx \showLCCN     \undefined \def \showLCCN      #1{\unskip}     \fi
\ifx \shownote     \undefined \def \shownote      #1{#1}          \fi
\ifx \showarticletitle \undefined \def \showarticletitle #1{#1}   \fi
\ifx \showURL      \undefined \def \showURL       {\relax}        \fi
% The following commands are used for tagged output and should be
% invisible to TeX
\providecommand\bibfield[2]{#2}
\providecommand\bibinfo[2]{#2}
\providecommand\natexlab[1]{#1}
\providecommand\showeprint[2][]{arXiv:#2}

\bibitem[\protect\citeauthoryear{??}{ms2}{2016}]%
        {ms2016}
 \bibinfo{year}{2016}\natexlab{}.
\newblock \bibinfo{title}{Microsoft Academic Graph - KDD cup 2016}.
\newblock
  \bibinfo{howpublished}{\url{https://kddcup2016.azurewebsites.net/Data}}.
  (\bibinfo{year}{2016}).
\newblock


\bibitem[\protect\citeauthoryear{??}{dbl}{2018}]%
        {dblp-ds2018}
 \bibinfo{year}{2018}\natexlab{}.
\newblock \bibinfo{title}{DBLP graphs}.
\newblock
  \bibinfo{howpublished}{\url{http://projects.csail.mit.edu/dnd/DBLP/}}.
  (\bibinfo{year}{2018}).
\newblock


\bibitem[\protect\citeauthoryear{??}{wik}{2018}]%
        {wiki-dataset}
 \bibinfo{year}{2018}\natexlab{}.
\newblock \bibinfo{title}{Deep Neural Networks Based Approaches for Graph
  Embeddings}.
\newblock   (\bibinfo{year}{2018}).
\newblock
\showURL{%
\url{https://github.com/PFE-Passau/Evaluation_Framework_For_Graph_Embedding}}


\bibitem[\protect\citeauthoryear{??}{net}{2018}]%
        {network-storage18}
 \bibinfo{year}{2018}\natexlab{}.
\newblock \bibinfo{title}{Massive network data}.
\newblock
  \bibinfo{howpublished}{\url{http://networkrepository.com/massive.php}}.
  (\bibinfo{year}{2018}).
\newblock


\bibitem[\protect\citeauthoryear{Abadi, Barham, Chen, Chen, Davis, Dean, Devin,
  Ghemawat, Irving, Isard, et~al\mbox{.}}{Abadi et~al\mbox{.}}{2016}]%
        {abadi2016tensorflow}
\bibfield{author}{\bibinfo{person}{Mart{\'\i}n Abadi}, \bibinfo{person}{Paul
  Barham}, \bibinfo{person}{Jianmin Chen}, \bibinfo{person}{Zhifeng Chen},
  \bibinfo{person}{Andy Davis}, \bibinfo{person}{Jeffrey Dean},
  \bibinfo{person}{Matthieu Devin}, \bibinfo{person}{Sanjay Ghemawat},
  \bibinfo{person}{Geoffrey Irving}, \bibinfo{person}{Michael Isard},
  {et~al\mbox{.}}} \bibinfo{year}{2016}\natexlab{}.
\newblock \showarticletitle{Tensorflow: a system for large-scale machine
  learning.}. In \bibinfo{booktitle}{{\em OSDI}}, Vol.~\bibinfo{volume}{16}.
  \bibinfo{pages}{265--283}.
\newblock


\bibitem[\protect\citeauthoryear{Adhikari, Zhang, Conference, and
  {2018}}{Adhikari et~al\mbox{.}}{[n. d.]}]%
        {adhikari2018sub2vec}
\bibfield{author}{\bibinfo{person}{B Adhikari}, \bibinfo{person}{Y Zhang},
  \bibinfo{person}{N~Ramakrishnan Pacific-Asia Conference}, {and}
  \bibinfo{person}{{2018}}.} \bibinfo{year}{[n. d.]}\natexlab{}.
\newblock \showarticletitle{{Sub2Vec: Feature Learning for Subgraphs}}.
\newblock \bibinfo{journal}{{\em Springer\/}} (\bibinfo{year}{[n. d.]}).
\newblock


\bibitem[\protect\citeauthoryear{Backstrom and Leskovec}{Backstrom and
  Leskovec}{2011}]%
        {backstrom2011supervised}
\bibfield{author}{\bibinfo{person}{Lars Backstrom} {and} \bibinfo{person}{Jure
  Leskovec}.} \bibinfo{year}{2011}\natexlab{}.
\newblock \showarticletitle{Supervised random walks: predicting and
  recommending links in social networks}. In \bibinfo{booktitle}{{\em
  Proceedings of the fourth ACM international conference on Web search and data
  mining}}. ACM, \bibinfo{pages}{635--644}.
\newblock


\bibitem[\protect\citeauthoryear{Belkin and Niyogi}{Belkin and Niyogi}{2002}]%
        {belkin2002laplacian}
\bibfield{author}{\bibinfo{person}{Mikhail Belkin} {and}
  \bibinfo{person}{Partha Niyogi}.} \bibinfo{year}{2002}\natexlab{}.
\newblock \showarticletitle{Laplacian eigenmaps and spectral techniques for
  embedding and clustering}. In \bibinfo{booktitle}{{\em Advances in neural
  information processing systems}}. \bibinfo{pages}{585--591}.
\newblock


\bibitem[\protect\citeauthoryear{Bengio}{Bengio}{2012}]%
        {bengio2012deep}
\bibfield{author}{\bibinfo{person}{Yoshua Bengio}.}
  \bibinfo{year}{2012}\natexlab{}.
\newblock \showarticletitle{Deep learning of representations for unsupervised
  and transfer learning}. In \bibinfo{booktitle}{{\em Proceedings of ICML
  Workshop on Unsupervised and Transfer Learning}}. \bibinfo{pages}{17--36}.
\newblock


\bibitem[\protect\citeauthoryear{Benson, Gleich, and Lim}{Benson
  et~al\mbox{.}}{2017}]%
        {benson2017spacey}
\bibfield{author}{\bibinfo{person}{Austin~R Benson}, \bibinfo{person}{David~F
  Gleich}, {and} \bibinfo{person}{Lek-Heng Lim}.}
  \bibinfo{year}{2017}\natexlab{}.
\newblock \showarticletitle{The spacey random walk: A stochastic process for
  higher-order data}.
\newblock \bibinfo{journal}{{\it SIAM Rev.}} \bibinfo{volume}{59},
  \bibinfo{number}{2} (\bibinfo{year}{2017}), \bibinfo{pages}{321--345}.
\newblock


\bibitem[\protect\citeauthoryear{Bollob{\'a}s}{Bollob{\'a}s}{1998}]%
        {bollobas1998graphtheory}
\bibfield{author}{\bibinfo{person}{B Bollob{\'a}s}.}
  \bibinfo{year}{1998}\natexlab{}.
\newblock \bibinfo{booktitle}{{\em {Modern Graph Theory}}}.
  Vol.~\bibinfo{volume}{Graduate Texts in Mathematics}.
\newblock


\bibitem[\protect\citeauthoryear{Chang, Zhang, Tang, Yin, Chang,
  Hasegawa-Johnson, and Huang}{Chang et~al\mbox{.}}{2017}]%
        {chang2017streaming}
\bibfield{author}{\bibinfo{person}{Shiyu Chang}, \bibinfo{person}{Yang Zhang},
  \bibinfo{person}{Jiliang Tang}, \bibinfo{person}{Dawei Yin},
  \bibinfo{person}{Yi Chang}, \bibinfo{person}{Mark~A Hasegawa-Johnson}, {and}
  \bibinfo{person}{Thomas~S Huang}.} \bibinfo{year}{2017}\natexlab{}.
\newblock \showarticletitle{Streaming recommender systems}. In
  \bibinfo{booktitle}{{\em Proceedings of the 26th International Conference on
  World Wide Web}}. International World Wide Web Conferences Steering
  Committee, \bibinfo{pages}{381--389}.
\newblock


\bibitem[\protect\citeauthoryear{Chen and Sun}{Chen and Sun}{2017}]%
        {chen2017task}
\bibfield{author}{\bibinfo{person}{Ting Chen} {and} \bibinfo{person}{Yizhou
  Sun}.} \bibinfo{year}{2017}\natexlab{}.
\newblock \showarticletitle{Task-guided and path-augmented heterogeneous
  network embedding for author identification}. In \bibinfo{booktitle}{{\em
  Proceedings of the Tenth ACM International Conference on Web Search and Data
  Mining}}. ACM, \bibinfo{pages}{295--304}.
\newblock


\bibitem[\protect\citeauthoryear{De~Winter, Decuypere, Mitrovic, Baesens, and
  De~Weerdt}{De~Winter et~al\mbox{.}}{[n. d.]}]%
        {dewinter2018combining}
\bibfield{author}{\bibinfo{person}{Sam De~Winter}, \bibinfo{person}{Tim
  Decuypere}, \bibinfo{person}{Sandra Mitrovic}, \bibinfo{person}{Bart
  Baesens}, {and} \bibinfo{person}{Jochen De~Weerdt}.} \bibinfo{year}{[n.
  d.]}\natexlab{}.
\newblock \showarticletitle{{Combining Temporal Aspects of Dynamic Networks
  with Node2Vec for a more Efficient Dynamic Link Prediction}}. In
  \bibinfo{booktitle}{{\em 2018 IEEE/ACM International Conference on Advances
  in Social Networks Analysis and Mining (ASONAM)}}. \bibinfo{publisher}{IEEE},
  \bibinfo{pages}{1234--1241}.
\newblock


\bibitem[\protect\citeauthoryear{Dong, Chawla, and Swami}{Dong
  et~al\mbox{.}}{2017}]%
        {dong2017metapath2vec}
\bibfield{author}{\bibinfo{person}{Yuxiao Dong}, \bibinfo{person}{Nitesh~V
  Chawla}, {and} \bibinfo{person}{Ananthram Swami}.}
  \bibinfo{year}{2017}\natexlab{}.
\newblock \showarticletitle{metapath2vec: Scalable representation learning for
  heterogeneous networks}. In \bibinfo{booktitle}{{\em Proceedings of the 23rd
  ACM SIGKDD International Conference on Knowledge Discovery and Data Mining}}.
  ACM, \bibinfo{pages}{135--144}.
\newblock


\bibitem[\protect\citeauthoryear{Du, Wang, Song, Lu, and Wang}{Du
  et~al\mbox{.}}{2018}]%
        {du2018dynamic}
\bibfield{author}{\bibinfo{person}{Lun Du}, \bibinfo{person}{Yun Wang},
  \bibinfo{person}{Guojie Song}, \bibinfo{person}{Zhicong Lu}, {and}
  \bibinfo{person}{Junshan Wang}.} \bibinfo{year}{2018}\natexlab{}.
\newblock \showarticletitle{Dynamic Network Embedding: An Extended Approach for
  Skip-gram based Network Embedding.}. In \bibinfo{booktitle}{{\em IJCAI}}.
  \bibinfo{pages}{2086--2092}.
\newblock


\bibitem[\protect\citeauthoryear{Grover and Leskovec}{Grover and
  Leskovec}{2016}]%
        {grover2016node2vec}
\bibfield{author}{\bibinfo{person}{Aditya Grover} {and} \bibinfo{person}{Jure
  Leskovec}.} \bibinfo{year}{2016}\natexlab{}.
\newblock \showarticletitle{node2vec: Scalable feature learning for networks}.
  In \bibinfo{booktitle}{{\em Proceedings of the 22nd ACM SIGKDD international
  conference on Knowledge discovery and data mining}}. ACM,
  \bibinfo{pages}{855--864}.
\newblock


\bibitem[\protect\citeauthoryear{Hamilton, Ying, and Leskovec}{Hamilton
  et~al\mbox{.}}{2017a}]%
        {hamilton2017inductive}
\bibfield{author}{\bibinfo{person}{Will Hamilton}, \bibinfo{person}{Zhitao
  Ying}, {and} \bibinfo{person}{Jure Leskovec}.}
  \bibinfo{year}{2017}\natexlab{a}.
\newblock \showarticletitle{Inductive representation learning on large graphs}.
  In \bibinfo{booktitle}{{\em Advances in Neural Information Processing
  Systems}}. \bibinfo{pages}{1025--1035}.
\newblock


\bibitem[\protect\citeauthoryear{Hamilton, Ying, and Leskovec}{Hamilton
  et~al\mbox{.}}{2017b}]%
        {hamilton2017representation}
\bibfield{author}{\bibinfo{person}{William~L Hamilton}, \bibinfo{person}{Rex
  Ying}, {and} \bibinfo{person}{Jure Leskovec}.}
  \bibinfo{year}{2017}\natexlab{b}.
\newblock \showarticletitle{{Representation Learning on Graphs: Methods and
  Applications}}.
\newblock \bibinfo{journal}{{\em arXiv.org\/}} (\bibinfo{date}{Sept.}
  \bibinfo{year}{2017}).
\newblock
\showeprint{1709.05584v3}


\bibitem[\protect\citeauthoryear{Henderson, Gallagher, Li, Akoglu, Eliassi-Rad,
  Tong, and Faloutsos}{Henderson et~al\mbox{.}}{2011}]%
        {henderson2011s}
\bibfield{author}{\bibinfo{person}{Keith Henderson}, \bibinfo{person}{Brian
  Gallagher}, \bibinfo{person}{Lei Li}, \bibinfo{person}{Leman Akoglu},
  \bibinfo{person}{Tina Eliassi-Rad}, \bibinfo{person}{Hanghang Tong}, {and}
  \bibinfo{person}{Christos Faloutsos}.} \bibinfo{year}{2011}\natexlab{}.
\newblock \showarticletitle{It's who you know: graph mining using recursive
  structural features}. In \bibinfo{booktitle}{{\em Proceedings of the 17th ACM
  SIGKDD international conference on Knowledge discovery and data mining}}.
  ACM, \bibinfo{pages}{663--671}.
\newblock


\bibitem[\protect\citeauthoryear{Kipf and Welling}{Kipf and Welling}{2016}]%
        {kipf2016semi}
\bibfield{author}{\bibinfo{person}{Thomas~N Kipf} {and} \bibinfo{person}{Max
  Welling}.} \bibinfo{year}{2016}\natexlab{}.
\newblock \showarticletitle{Semi-supervised classification with graph
  convolutional networks}.
\newblock \bibinfo{journal}{{\em arXiv preprint arXiv:1609.02907\/}}
  (\bibinfo{year}{2016}).
\newblock


\bibitem[\protect\citeauthoryear{Li, Cheng, Wu, and Liu}{Li
  et~al\mbox{.}}{2018}]%
        {li2018streaming}
\bibfield{author}{\bibinfo{person}{Jundong Li}, \bibinfo{person}{Kewei Cheng},
  \bibinfo{person}{Liang Wu}, {and} \bibinfo{person}{Huan Liu}.}
  \bibinfo{year}{2018}\natexlab{}.
\newblock \showarticletitle{Streaming link prediction on dynamic attributed
  networks}. In \bibinfo{booktitle}{{\em Proceedings of the Eleventh ACM
  International Conference on Web Search and Data Mining}}. ACM,
  \bibinfo{pages}{369--377}.
\newblock


\bibitem[\protect\citeauthoryear{Li, Dani, Hu, Tang, Chang, and Liu}{Li
  et~al\mbox{.}}{2017}]%
        {li2017attributed}
\bibfield{author}{\bibinfo{person}{Jundong Li}, \bibinfo{person}{Harsh Dani},
  \bibinfo{person}{Xia Hu}, \bibinfo{person}{Jiliang Tang}, \bibinfo{person}{Yi
  Chang}, {and} \bibinfo{person}{Huan Liu}.} \bibinfo{year}{2017}\natexlab{}.
\newblock \showarticletitle{Attributed network embedding for learning in a
  dynamic environment}. In \bibinfo{booktitle}{{\em Proceedings of the 2017 ACM
  on Conference on Information and Knowledge Management}}. ACM,
  \bibinfo{pages}{387--396}.
\newblock


\bibitem[\protect\citeauthoryear{McCallum, Nigam, Rennie, and Seymore}{McCallum
  et~al\mbox{.}}{2000}]%
        {mccallum2000automating}
\bibfield{author}{\bibinfo{person}{Andrew~Kachites McCallum},
  \bibinfo{person}{Kamal Nigam}, \bibinfo{person}{Jason Rennie}, {and}
  \bibinfo{person}{Kristie Seymore}.} \bibinfo{year}{2000}\natexlab{}.
\newblock \showarticletitle{Automating the construction of internet portals
  with machine learning}.
\newblock \bibinfo{journal}{{\em Information Retrieval\/}} \bibinfo{volume}{3},
  \bibinfo{number}{2} (\bibinfo{year}{2000}), \bibinfo{pages}{127--163}.
\newblock


\bibitem[\protect\citeauthoryear{Mikolov, Chen, Corrado, and Dean}{Mikolov
  et~al\mbox{.}}{2013a}]%
        {mikolov2013efficient}
\bibfield{author}{\bibinfo{person}{Tomas Mikolov}, \bibinfo{person}{Kai Chen},
  \bibinfo{person}{Greg Corrado}, {and} \bibinfo{person}{Jeffrey Dean}.}
  \bibinfo{year}{2013}\natexlab{a}.
\newblock \showarticletitle{Efficient estimation of word representations in
  vector space}.
\newblock \bibinfo{journal}{{\em arXiv preprint arXiv:1301.3781\/}}
  (\bibinfo{year}{2013}).
\newblock


\bibitem[\protect\citeauthoryear{Mikolov, Sutskever, Chen, Corrado, and
  Dean}{Mikolov et~al\mbox{.}}{2013b}]%
        {mikolov2013distributed}
\bibfield{author}{\bibinfo{person}{Tomas Mikolov}, \bibinfo{person}{Ilya
  Sutskever}, \bibinfo{person}{Kai Chen}, \bibinfo{person}{Greg~S Corrado},
  {and} \bibinfo{person}{Jeff Dean}.} \bibinfo{year}{2013}\natexlab{b}.
\newblock \showarticletitle{Distributed representations of words and phrases
  and their compositionality}. In \bibinfo{booktitle}{{\em Advances in neural
  information processing systems}}. \bibinfo{pages}{3111--3119}.
\newblock


\bibitem[\protect\citeauthoryear{Nguyen, Lee, Rossi, Ahmed, Koh, and
  Kim}{Nguyen et~al\mbox{.}}{2018a}]%
        {nguyen2018continuous}
\bibfield{author}{\bibinfo{person}{Giang~Hoang Nguyen},
  \bibinfo{person}{John~Boaz Lee}, \bibinfo{person}{Ryan~A Rossi},
  \bibinfo{person}{Nesreen~K Ahmed}, \bibinfo{person}{Eunyee Koh}, {and}
  \bibinfo{person}{Sungchul Kim}.} \bibinfo{year}{2018}\natexlab{a}.
\newblock \showarticletitle{Continuous-time dynamic network embeddings}. In
  \bibinfo{booktitle}{{\em 3rd International Workshop on Learning
  Representations for Big Networks (WWW BigNet)}}.
\newblock


\bibitem[\protect\citeauthoryear{Nguyen, Lee, Rossi, Ahmed, Koh, and
  Kim}{Nguyen et~al\mbox{.}}{2018b}]%
        {nguyen2018dynamic}
\bibfield{author}{\bibinfo{person}{Giang~Hoang Nguyen},
  \bibinfo{person}{John~Boaz Lee}, \bibinfo{person}{Ryan~A. Rossi},
  \bibinfo{person}{Nesreen~K. Ahmed}, \bibinfo{person}{Eunyee Koh}, {and}
  \bibinfo{person}{Sungchul Kim}.} \bibinfo{year}{2018}\natexlab{b}.
\newblock \showarticletitle{{Dynamic Network Embeddings: From Random Walks to
  Temporal Random Walks}}. In \bibinfo{booktitle}{{\em IEEE BigData}}.
\newblock


\bibitem[\protect\citeauthoryear{Perozzi, Al-Rfou, and Skiena}{Perozzi
  et~al\mbox{.}}{2014}]%
        {perozzi2014deepwalk}
\bibfield{author}{\bibinfo{person}{Bryan Perozzi}, \bibinfo{person}{Rami
  Al-Rfou}, {and} \bibinfo{person}{Steven Skiena}.}
  \bibinfo{year}{2014}\natexlab{}.
\newblock \showarticletitle{Deepwalk: Online learning of social
  representations}. In \bibinfo{booktitle}{{\em Proceedings of the 20th ACM
  SIGKDD international conference on Knowledge discovery and data mining}}.
  ACM, \bibinfo{pages}{701--710}.
\newblock


\bibitem[\protect\citeauthoryear{Qiu, Dong, Ma, Li, Wang, and Tang}{Qiu
  et~al\mbox{.}}{2018}]%
        {qiu2018network}
\bibfield{author}{\bibinfo{person}{Jiezhong Qiu}, \bibinfo{person}{Yuxiao
  Dong}, \bibinfo{person}{Hao Ma}, \bibinfo{person}{Jian Li},
  \bibinfo{person}{Kuansan Wang}, {and} \bibinfo{person}{Jie Tang}.}
  \bibinfo{year}{2018}\natexlab{}.
\newblock \showarticletitle{Network Embedding as Matrix Factorization: Unifying
  DeepWalk, LINE, PTE, and node2vec}. In \bibinfo{booktitle}{{\em Proceedings
  of the Eleventh ACM International Conference on Web Search and Data Mining}}.
  ACM, \bibinfo{pages}{459--467}.
\newblock


\bibitem[\protect\citeauthoryear{Ravi and Larochelle}{Ravi and
  Larochelle}{2016}]%
        {ravi2016optimization}
\bibfield{author}{\bibinfo{person}{Sachin Ravi} {and} \bibinfo{person}{Hugo
  Larochelle}.} \bibinfo{year}{2016}\natexlab{}.
\newblock \showarticletitle{Optimization as a model for few-shot learning}.
\newblock  (\bibinfo{year}{2016}).
\newblock


\bibitem[\protect\citeauthoryear{Roweis and Saul}{Roweis and Saul}{2000}]%
        {roweis2000nonlinear}
\bibfield{author}{\bibinfo{person}{Sam~T Roweis} {and}
  \bibinfo{person}{Lawrence~K Saul}.} \bibinfo{year}{2000}\natexlab{}.
\newblock \showarticletitle{Nonlinear dimensionality reduction by locally
  linear embedding}.
\newblock \bibinfo{journal}{{\em science\/}} \bibinfo{volume}{290},
  \bibinfo{number}{5500} (\bibinfo{year}{2000}), \bibinfo{pages}{2323--2326}.
\newblock


\bibitem[\protect\citeauthoryear{Tang and Liu}{Tang and Liu}{2012}]%
        {tang2012unsupervised}
\bibfield{author}{\bibinfo{person}{Jiliang Tang} {and} \bibinfo{person}{Huan
  Liu}.} \bibinfo{year}{2012}\natexlab{}.
\newblock \showarticletitle{Unsupervised feature selection for linked social
  media data}. In \bibinfo{booktitle}{{\em Proceedings of the 18th ACM SIGKDD
  international conference on Knowledge discovery and data mining}}. ACM,
  \bibinfo{pages}{904--912}.
\newblock


\bibitem[\protect\citeauthoryear{Tang, Qu, Wang, Zhang, Yan, and Mei}{Tang
  et~al\mbox{.}}{2015}]%
        {tang2015line}
\bibfield{author}{\bibinfo{person}{Jian Tang}, \bibinfo{person}{Meng Qu},
  \bibinfo{person}{Mingzhe Wang}, \bibinfo{person}{Ming Zhang},
  \bibinfo{person}{Jun Yan}, {and} \bibinfo{person}{Qiaozhu Mei}.}
  \bibinfo{year}{2015}\natexlab{}.
\newblock \showarticletitle{Line: Large-scale information network embedding}.
  In \bibinfo{booktitle}{{\em Proceedings of the 24th International Conference
  on World Wide Web}}. International World Wide Web Conferences Steering
  Committee, \bibinfo{pages}{1067--1077}.
\newblock


\bibitem[\protect\citeauthoryear{Tenenbaum, De~Silva, and Langford}{Tenenbaum
  et~al\mbox{.}}{2000}]%
        {tenenbaum2000global}
\bibfield{author}{\bibinfo{person}{Joshua~B Tenenbaum}, \bibinfo{person}{Vin
  De~Silva}, {and} \bibinfo{person}{John~C Langford}.}
  \bibinfo{year}{2000}\natexlab{}.
\newblock \showarticletitle{A global geometric framework for nonlinear
  dimensionality reduction}.
\newblock \bibinfo{journal}{{\em science\/}} \bibinfo{volume}{290},
  \bibinfo{number}{5500} (\bibinfo{year}{2000}), \bibinfo{pages}{2319--2323}.
\newblock


\bibitem[\protect\citeauthoryear{Tsitsulin, Mottin, Karras, and
  M{\"u}ller}{Tsitsulin et~al\mbox{.}}{2018}]%
        {tsitsulin2018verse}
\bibfield{author}{\bibinfo{person}{Anton Tsitsulin}, \bibinfo{person}{Davide
  Mottin}, \bibinfo{person}{Panagiotis Karras}, {and} \bibinfo{person}{Emmanuel
  M{\"u}ller}.} \bibinfo{year}{2018}\natexlab{}.
\newblock \showarticletitle{VERSE: Versatile Graph Embeddings from Similarity
  Measures}.
\newblock  (\bibinfo{year}{2018}).
\newblock


\bibitem[\protect\citeauthoryear{Tsoumakas and Katakis}{Tsoumakas and
  Katakis}{2007}]%
        {tsoumakas2007multi}
\bibfield{author}{\bibinfo{person}{Grigorios Tsoumakas} {and}
  \bibinfo{person}{Ioannis Katakis}.} \bibinfo{year}{2007}\natexlab{}.
\newblock \showarticletitle{Multi-label classification: An overview}.
\newblock \bibinfo{journal}{{\em International Journal of Data Warehousing and
  Mining (IJDWM)\/}} \bibinfo{volume}{3}, \bibinfo{number}{3}
  (\bibinfo{year}{2007}), \bibinfo{pages}{1--13}.
\newblock


\bibitem[\protect\citeauthoryear{Ying, He, Chen, Eksombatchai, Hamilton, and
  Leskovec}{Ying et~al\mbox{.}}{2018}]%
        {ying2018graph}
\bibfield{author}{\bibinfo{person}{Rex Ying}, \bibinfo{person}{Ruining He},
  \bibinfo{person}{Kaifeng Chen}, \bibinfo{person}{Pong Eksombatchai},
  \bibinfo{person}{William~L Hamilton}, {and} \bibinfo{person}{Jure Leskovec}.}
  \bibinfo{year}{2018}\natexlab{}.
\newblock \showarticletitle{Graph Convolutional Neural Networks for Web-Scale
  Recommender Systems}.
\newblock \bibinfo{journal}{{\em arXiv preprint arXiv:1806.01973\/}}
  (\bibinfo{year}{2018}).
\newblock


\bibitem[\protect\citeauthoryear{Zafarani and Liu}{Zafarani and Liu}{2009}]%
        {zafarani2009social}
\bibfield{author}{\bibinfo{person}{Reza Zafarani} {and} \bibinfo{person}{Huan
  Liu}.} \bibinfo{year}{2009}\natexlab{}.
\newblock \bibinfo{title}{Social computing data repository at ASU}.
\newblock   (\bibinfo{year}{2009}).
\newblock


\bibitem[\protect\citeauthoryear{Zaharia, Chowdhury, Franklin, Shenker, and
  Stoica}{Zaharia et~al\mbox{.}}{2010}]%
        {zaharia2010spark}
\bibfield{author}{\bibinfo{person}{Matei Zaharia}, \bibinfo{person}{Mosharaf
  Chowdhury}, \bibinfo{person}{Michael~J Franklin}, \bibinfo{person}{Scott
  Shenker}, {and} \bibinfo{person}{Ion Stoica}.}
  \bibinfo{year}{2010}\natexlab{}.
\newblock \showarticletitle{Spark: Cluster computing with working sets.}
\newblock \bibinfo{journal}{{\em HotCloud\/}} \bibinfo{volume}{10},
  \bibinfo{number}{10-10} (\bibinfo{year}{2010}), \bibinfo{pages}{95}.
\newblock


\bibitem[\protect\citeauthoryear{Zhang, Yin, Zhu, and Zhang}{Zhang
  et~al\mbox{.}}{2018}]%
        {zhang2018review}
\bibfield{author}{\bibinfo{person}{Daokun Zhang}, \bibinfo{person}{Jie Yin},
  \bibinfo{person}{Xingquan Zhu}, {and} \bibinfo{person}{Chengqi Zhang}.}
  \bibinfo{year}{2018}\natexlab{}.
\newblock \showarticletitle{Network representation learning: a survey}.
\newblock \bibinfo{journal}{{\em IEEE Transactions on Knowledge and Data
  Engineering\/}}.
\newblock
\showISSN{2332-7790}
\showDOI{%
\url{https://doi.org/10.1109/TBDATA.2018.2850013}}


\end{thebibliography}
